\documentclass{article}

\PassOptionsToPackage{numbers, compress}{natbib}

\usepackage[preprint]{neurips_2022}

\usepackage[utf8]{inputenc} 
\usepackage[T1]{fontenc}    
\usepackage{hyperref}       
\usepackage{url}            
\usepackage{booktabs}       
\usepackage{amsfonts}       
\usepackage{nicefrac}       
\usepackage{microtype}      
\usepackage{xcolor}         
\usepackage{pifont}         
\usepackage{amsmath,amssymb,amsfonts,bm}
\usepackage{graphicx}
\usepackage{caption}
\usepackage{subcaption}
\usepackage{algorithmic}
\usepackage[ruled,vlined]{algorithm2e}

\title{On the Relationship Between Variational Inference and Auto-Associative Memory}

\author{%
  Louis Annabi \\
  ETIS UMR 8051\\
  CY Cergy Paris Université, ENSEA, CNRS\\
  Cergy, France \\
  \texttt{louis.annabi@gmail.com} \\
  \And
  Alexandre Pitti \\
  ETIS UMR 8051\\
  CY Cergy Paris Université, ENSEA, CNRS\\
  Cergy, France \\
  \texttt{alexandre.pitti@ensea.fr} \\
  \AND
  Mathias Quoy \\
  ETIS UMR 8051\\
  CY Cergy Paris Université, ENSEA, CNRS\\
  Cergy, France \\
  \texttt{mathias.quoy@ensea.fr} \\
}

\begin{document}

\maketitle

\begin{abstract}

In this article, we propose a variational inference formulation of auto-associative memories, allowing us to combine perceptual inference and memory retrieval into the same mathematical framework. In this formulation, the prior probability distribution onto latent representations is made memory dependent, thus pulling the inference process towards previously stored representations. We then study how different neural network approaches to variational inference can be applied in this framework. We compare methods relying on amortized inference such as Variational Auto Encoders and methods relying on iterative inference such as Predictive Coding and suggest combining both approaches to design new auto-associative memory models. We evaluate the obtained algorithms on the CIFAR10 and CLEVR image datasets and compare them with other associative memory models such as Hopfield Networks, End-to-End Memory Networks and Neural Turing Machines.

\end{abstract}

\section{Introduction}

In the recent years, many methods have been proposed in order to augment deep learning models with long-term memories. These models provide writing and reading mechanisms allowing to store and retrieve certain patterns inside a memory. Among these models, associative memories are memories that perform content-based addressing, meaning that they retrieve stored patterns based on an approximate or incomplete version of those.

While many models consider one-step reading mechanisms to retrieve patterns from the memory, biological memory retrieval is an iterative process that can take a variable amount of time. Taking this into consideration, Hopfield networks \citep{Hopfield1982} have been proposed as an Associative Memory (AM) implemented by a Recurrent Neural Network (RNN). Hopfield networks perform memory retrieval as an iterative process where stored patterns constitute local attractors of the RNN dynamics.

In this work, we show how AM models such as Hopfield networks can be formulated as variational inference methods using a memory dependent probabilistic model of the observed data. In variational inference algorithms such as Variational Auto-Encoders (VAE) \citep{Kingma2014, Rezende2014} and Predictive Coding (PC) \citep{Rao1999, Friston2009c}, the representation inferred from the observed data tries to minimize an energy function based on the probabilistic model $p(\bm{x})$ generating the observed data. In the Free Energy Principle (FEP) literature \citep{Friston2009b, Friston2010}, this function is called Variational Free-Energy (VFE), and is equivalent to the negative Evidence Lower Bound (ELBO) more often used in machine learning. The probabilistic model $p(\bm{x})$ can be decomposed according to a prior probability distribution over representations $p(\bm{z})$, and a likelihood $p(\bm{x}|\bm{z})$ describing the probability of observed data based on the representation. We propose to make $p(\bm{z})$ depend on the patterns stored in memory, which allows us to derive a new expression of the VFE. The obtained energy function has minima that should be close to one of the stored patterns while properly encoding the observed input. As such, inference based on this energy function can be seen both as perceptual inference and memory retrieval. While using memory-dependent generative models or Gaussian Mixture Models (GMM) with VAEs \citep{dilokthanakul2016deep, bornschein2017variational, le2018variational} are not novel ideas, the PC approach has never been to this problem.

The paper is organized as follows: in section \ref{sec:related_work}, we provide a deeper presentation of the concepts connected to our work and review related approaches. In section \ref{sec:theory} we present an overview of the FEP mathematical framework and derive an expression of the VFE depending on the patterns stored in memory. In section \ref{sec:methods} we design several AM models minimizing this energy function. In section \ref{sec:experiments}, we evaluate the obtained algorithms on the task of memory retrieval on two image datasets, and compare their performance with other AM models.

The contributions brought by this work are the following:
\begin{itemize}
    \item We design four AM models based on a variational inference formulation of memory retrieval.
    \item We draw a connection between PC and the modern continuous Hopfield network \citep{Ramsauer2020}.
\end{itemize}

\section{Related work}
\label{sec:related_work}

\subsection{Auto-associative memory}

Several approaches to the long-term storage of information in artificial neural networks have been proposed. Memory networks \citep{Weston2015}, and End-to-End Memory Networks (MemN2N) \citep{Sukhbaatar2015} propose to store information in a memory matrix that can be addressed using attention coefficients computed based on the content of different memory locations. The Neural Turing Machine (NTM) \citep{Graves2014} and the differentiable neural computer \citep{Graves2016} combine this content-based addressing using attention mechanisms with a location-based addressing allowing more computer-like memory accesses.

Standing out from these approaches that consider a feedforward reading mechanism, Hopfield networks \citep{Hopfield1982} and continuous Hopfield networks \citep{Hopfield1984} instead store patterns as local attractors of an RNN. These memories can thus be addressed by initializing the RNN state with the input pattern, and retrieving the stored attractor after convergence. In order to improve the memory capacity, modern Hopfield networks \citep{Krotov2016, Krotov2018, Demircigil2017} propose several variants of the energy function using polynomial or exponential interactions. Extending these models to the continuous case, \cite{Ramsauer2020} proposed the Modern Continuous Hopfield Network (MCHN) with update rules implementing self attention, that they relate to the transformer model \citep{Vaswani2017}. In \cite{Millidge2022}, the authors introduce a general Hopfield network framework where the update rules are built using three components: a similarity function, a separation function, and a projection function.

It has been shown that overparameterized auto-encoders also implement AM \citep{radhakrishnan2020overparameterized, Salvatori2021}. These methods embed the stored patterns as attractors through training, and retrieval is performed by iterating over the auto-encoding loop. In contrast, our methods allow one-shot writing: a new pattern $\bm{x}$ can be inserted in memory simply by computing its representation and adding it to the memory matrix $\bm{M}$.

Another line of research takes inspiration from the Sparse Distributed Memory model \citep{kanerva1988sparse}, building connections with attention mechanisms \citep{bricken2021attention} and with the variational inference framework \citep{wu2018kanerva, wu2018learning, ramapuram2020kanerva++}. In particular, the Kanerva Machine \citep{wu2018kanerva, wu2018learning, ramapuram2020kanerva++} is similar to the models we build in many aspects: they use an iterative reading mechanism, a memory-dependent prior on the representation, all within the variational inference framework. Though, the PC toolbox provides methods that set the proposed models apart from these approaches, as detailed in the methods section.

\subsection{Predictive Coding}

PC is a theory of brain function \cite{Rao1999, Clark2013} extending the idea that neural representations emerge as part of an inference process of the causes of sensory observations, as already suggested by Helmholtz in 1867 \cite{Helmholtz1867}. The FEP \cite{Friston2009b, Friston2010} provides a principled derivation of PC networks based on the minimization of variational free-energy (VFE), a quantity equivalent to the negative evidence lower bound (ELBO) used in variational Bayesian methods, and defined as:

\begin{equation}
    F(\bm{x}) = \int_z \log \Big( \frac{q(\bm{z})}{p(\bm{x},\bm{z})} \Big) q(\bm{z}) \bm{dz}
    \label{eq:vfe}
\end{equation}

where $p(\bm{x}, \bm{z})$ denotes the generative model and $q(\bm{z})$ denotes the approximate posterior (also called recognition density) on $\bm{z}$. VAEs \cite{Kingma2014, Rezende2014} are a well-known method applying the idea of VFE minimization (equivalently ELBO maximization) to neural networks. In VAEs, this quantity is only optimized during the model training, and perceptual inference is performed as a simple forward pass through the encoder. 

In contrast, neural network models based on PC intertwine the prediction (decoder) and inference (encoder) mechanisms within a single hierarchical recurrent architecture comprising a population of representation neurons and a population of prediction error neurons at each layer. Perceptual inference is then an iterative mechanism supported by the dynamics of this RNN. Given an observed input $\bm{x}$, the representation $\bm{z}$ is updated at each time step based on a bottom-up signal pushing $\bm{z}$ towards values that minimize the reconstruction error, and on a top-down signal pushing $\bm{z}$ towards values that maximize its prior probability $p(\bm{z})$. While this prior probability on $\bm{z}$ is often ignored, we show that it can bring auto-associative capacities to the PC network, and we draw a connection between the obtained mechanism and the modern continuous Hopfield network proposed in \cite{Ramsauer2020}.


While both VAEs and PC networks output an estimation of the approximate posterior $q(\bm{z})$, they differ in the computational mechanisms used for inference: VAEs perform amortized inference via a forward pass through an encoder, while PC networks perform an iterative inference implemented by its recurrent dynamics. Some works have suggested combining both approaches: the iterative inference can be initialized using the estimation provided by the amortized inference method \citep{Bengio2016, Tschantz2022}. 

A previous work \citep{Annabi2021} explored the use PC techniques together with GMM-based prior probability distributions $p(\bm{z})$. However, the obtained models were not used to perform memory retrieval.

\section{Memory-based Variational Inference}
\label{sec:theory}

\subsection{Framework definition}

Our framework is based on a generative model $p(\bm{x}, \bm{z})$ that can be factored into a prior probability $p(\bm{z}$) and a likelihood $p(\bm{x}|\bm{z})$. The prior probability on $\bm{z}$ is defined as a memory dependent distribution:
\begin{equation}
    p(\bm{z}) = p(\bm{z}; \bm{M})
\end{equation}
where the vectors $\bm{M}_k$ (the columns of $\bm{M}$) constitute a repertoire of stored representations.

The probability $p(\bm{x}|\bm{z})$ is defined as a hierarchical generative model featuring several intermediate variables $\{\bm{h}_1, \cdots, \bm{h}_{L-1} \}$, where $L$ denotes the number of layers. By extension we use the notations $\bm{h}_0 = \bm{x}$ and $\bm{h}_L = \bm{z}$. We assume that the generative model is a cascade of multivariate Gaussians:
\begin{equation}
    p(\bm{h}_l|\bm{h}_{l+1}) = \mathcal{N}(\bm{h}_l;\bm{f_\theta}^{l}(\bm{h}_{l+1}), \mathbb{I})
\end{equation}
where  $\mathbb{I}$ is the covariance matrix of the Gaussians and is uniform across all layers, and $\bm{f_\theta}^l$ are functions (typically neural network layers) parameterized by $\bm{\theta}$. Note that this can be adapted to arbitrary computation graphs by replacing $\bm{h}_{l+1}$ by the set of parent nodes of $\bm{h}_l$ for each node. For simplicity, we assume in the following derivations that each node only has one parent. In our experiments, the functions $\bm{f_\theta}^{l}$ correspond to the different layers of a Convolutional Neural Network (CNN) on the CIFAR10 dataset, and of a MONet \citep{Burgess2019} decoder on the CLEVR dataset. We denote by $\bm{f_\theta}$ the composition of all layers $\bm{f_\theta}^l$.

We can derive the VFE corresponding to the described generative model. The FEP formulation of PC uses different approximations that allow us to greatly simplify the expression of the VFE introduced in equation (\ref{eq:vfe}). We refer to appendix \ref{app:fe_derivations} for detailed derivations and simply provide the simplified expression:

\begin{equation}
\label{eq:vfe_pc_expression}
    F(\bm{x}, \hat{\bm{h}_1}, \dots, \hat{\bm{h}_{L-1}}, \hat{\bm{z}}) = \sum_{l=0}^{L-1} \frac{1}{2} \| \hat{\bm{h}_l} - \bm{f_\theta}^{l}(\hat{\bm{h}_{l+1}}) \|^2
    - \log p(\hat{\bm{z}};\bm{M})
    + C
\end{equation}

where C is a quantity independent from $\{\bm{x}, \hat{\bm{h}_1}, \dots, \hat{\bm{h}_{L-1}}, \hat{\bm{z}}\}$. The vectors $\{\hat{\bm{h}_1}, \dots, \hat{\bm{h}_{L-1}}, \hat{\bm{z}}\}$ correspond to the means of the approximate posterior $q(\bm{h}_1, \dots, \bm{h}_{L-1}, \bm{z}) = q(\bm{z}) \prod_{i=1}^{L-1} q(\bm{h_i})$.

The term corresponding to the prior probability is often omitted, which can be justified as being equivalent to having no prior preferences over different values of $\bm{z}$. Key to our method is the idea that this prior can pull the inference process towards values of $\bm{z}$ previously stored in the memory $\bm{M}$, turning the patterns $\bm{M}_k$ into attractors of the PC network. As such, a suitable distribution would be one that associates high probabilities for the patterns stored in memory.

\subsection{Classification of related methods in this framework}
\label{sec:pc_mchn}

Here, we show that we can formulate Modern Continuous Hopfield Networks (MCHN) as PC networks derived from this expression of the VFE under some conditions:
\begin{itemize}
    \item There is no representation component: $L=0$.
    \item The PC network is initialized with $\hat{\bm{z}} = \bm{x}$.
    \item The prior distribution $p(\bm{z}; \bm{M})$ is defined as:
\end{itemize}

    \begin{align}
        p_{MCHN}(\bm{z};\bm{M}) &= \sum_{k=1}^N  \pi_k \mathcal{N}\big(\bm{z}; \bm{M}_k, \beta^{-1}\mathbb{I}\big) \\
        \text{with }\forall k, \pi_k &= \frac{\exp\{\frac{\beta}{2} \bm{M}_k^\intercal \cdot \bm{M}_k\}}{\sum_{k'=1}^N \exp\{\frac{\beta}{2} \bm{M}_{k'}^\intercal \cdot \bm{M}_{k'}\}}
    \end{align}
    
This prior distribution is a Gaussian Mixture Model (GMM) with mixture means corresponding to the $N$ stored patterns and mixing coefficients $\pi_k$ depending on the patterns' Euclidean norm. It is parameterized by a coefficient $\beta>0$. Based on these assumptions, we can derive the following expression for the VFE:

\begin{equation}
\label{eq:vfe_hopfield}
    F(\hat{\bm{z}}) = \frac{\beta}{2} \hat{\bm{z}}^\intercal \cdot \hat{\bm{z}} - \log \sum_{k=1}^N \exp\{\beta \hat{\bm{z}}^\intercal \cdot \bm{M}_k\} + C
\end{equation}

which is, up to a constant $C$ and a factor $\beta$, equivalent to the MCHN energy function proposed in \citep{Ramsauer2020}. The complete derivations of this expression are provided in appendix \ref{app:mchn_derivations}. We can note that VFE does not depend on the input $\bm{x}$. The input is not part of the energy function, but serves as an initial estimate of the approximate posterior mean $\hat{\bm{z}}$. According to the FEP formulation of PC, applying gradient descent on the VFE with regard to the approximate posterior mean $\hat{\bm{z}}$ yields the update rules of the PC network. Using our expression of the VFE, we obtain:

\begin{equation}
\label{eq:vfe_hopfield_update}
    \hat{\bm{z}} \gets \text{softmax}(\beta \hat{\bm{z}} \cdot \bm{M}) \bm{M}^\intercal
\end{equation}

This equation is identical to the update rule of MCHN (detailed derivations are provided in appendix). We have shown that our framework can be related to MCHNs, but variational inference methods for perceptual inference can also be retrieved in this framework by simply using a neutral prior. 

For VAEs and variants, this prior probability distribution is exploited during learning, but the amortized inference cannot be dynamically adapted to this distribution. Therefore, to take into account a change in this prior, we would need to retrain the encoder. In the derivations of PC networks \citep{Friston2009c, Buckley2017}, the term $p(\bm{z})$ is often ignored, which is equivalent to having a flat prior distribution. However, theoretically the iterative inference could take into account a prior distribution, and dynamically adapt to changes in this distribution, without retraining the model.

In the family of iterative inference algorithms, we can also mention the methods based on backpropagation (BP) to estimate the representation. Instead of using neural computations to simulate the gradient descent on the energy function as done in PC, these methods directly use BP to optimize the representation $\hat{\bm{z}}$. For instance \cite{Otte2017} minimize prediction error using BPTT to adjust a latent variable in an RNN. This method is in fact very similar to PC, since it has been shown that under some conditions, PC approximates the update rules entailed by BP applied on the reconstruction error \cite{Whittington2017, Millidge2020}. We provide in table \ref{tab:models} a simple classification of these approaches depending on whether they use a representation and memory component, and whether the inference mechanism is amortized or iterative.

\begin{table}
  \caption{Summary of the related models and proposed models.}
  \label{tab:models}
  \centering
  \begin{tabular}{lcc}
    \toprule
    Model                       & Representation            & Associative memory    \\
    \midrule
    MCHN \citep{Ramsauer2020}   & \ding{55}                 & Iterative             \\
    VAE \citep{Kingma2014}      & Amortized         & \ding{55}             \\
    PC  \citep{Rao1999}         & Iterative                 & \ding{55}             \\
    BP  \citep{Otte2017}        & Iterative                 & \ding{55}             \\
    HPC \citep{Tschantz2022}    & Amortized and iterative   & \ding{55}             \\
    GMVAE \citep{dilokthanakul2016deep}      & Amortized         & Amortized            \\
    Overparameterized VAE \citep{radhakrishnan2020overparameterized}      & Amortized         & Iterative            \\
    Kanerva Machines \citep{wu2018kanerva}      & Amortized         & Iterative            \\
    \midrule
    VAE-PC-GMM (Ours)           & Amortized and iterative   & Iterative             \\
    VAE-BP-GMM (Ours)           & Amortized and iterative   & Iterative             \\
    VAE-GMM (Ours)              & Amortized                 & Iterative             \\
    \bottomrule
  \end{tabular}
\end{table}

\section{Methods}
\label{sec:methods}

\begin{figure}[ht!]
    \centering
    \includegraphics[width=0.9\textwidth]{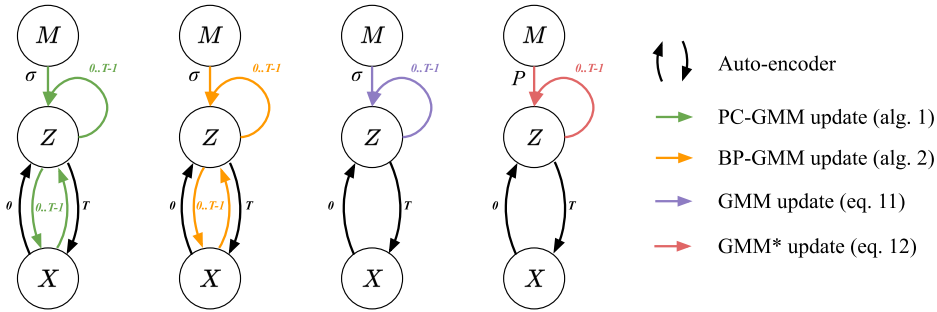}
    \caption{Representation of the memory retrieval process in the different proposed models. From left to right: VAE-PC-GMM (\ref{sec:vae_pc_gmm}), VAE-BP-GMM (\ref{sec:vae_bp_gmm}), VAE-GMM (\ref{sec:vae_gmm}) and VAE-GMM* (\ref{sec:vae_gmm_star}).}
    \label{fig:proposed_models}
\end{figure}

In this section, we present the AM models that we have designed using this framework. All these models are based on a pre-trained VAE that is used to provide initial estimates $\tilde{\bm{z}}$ and predictions of $\bm{x}$ based on $\tilde{\bm{z}}$. The proposed methods are represented in figure \ref{fig:proposed_models}. 

\subsection{Probabilistic model}

The prior probability over the representation $\bm{z}$ is crucial to obtain a model that infers representations close to stored patterns. We have shown that MCHN can be obtained starting from a GMM prior biased towards stored patterns of large Euclidean norm. Since there is no intuitive justification for this bias, we choose to instead start with a balanced GMM prior distribution:

\begin{equation}
    p(\bm{z};\bm{M}) = \sum_{k=1}^N \frac{1}{N} \mathcal{N}(\bm{z}; \bm{M}_k, \Sigma)
\end{equation}

where $\Sigma$ denotes the covariance matrix, uniform across the $N$ mixtures. In the following models, we consider the simpler case where $\bm{\Sigma} = \sigma^2 \mathbb{I}$, except for the last proposed model (section \ref{sec:vae_gmm_star}) where the precision matrix $\bm{P} = \bm{\Sigma}^{-1}$ is trained.

\subsection{PC based inference (VAE-PC-GMM)}
\label{sec:vae_pc_gmm}

Following the FEP formulation of PC, we can derive the gradient of $F$ (from eq. \ref{eq:vfe_pc_expression}) according to each layer's representation $\hat{\bm{h}_l}$. As such, performing gradient descent on $F$ yields a system of update rules on each representation, that can be interpreted as RNN dynamics. This PC network comprises at each layer two populations of neurons, one encoding the layer's representation, and one encoding the layer prediction error $\bm{\epsilon}_l = \hat{\bm{h}_l} - \bm{f_\theta}^{l}(\hat{\bm{h}_{l+1}})$. 

Detailed derivations of this model are provided in appendix \ref{app:pc_model}, where the forward pass through the model is given in algorithm \ref{alg:perceptual_inference}. Since the RNN dynamics implement a gradient descent, the initialization of this network is responsible for the local minimum to which it converges. We label VAE-PC-GMM the version of this algorithm where $\hat{\bm{z}}$ is initialized using the result of the amortized inference via the encoder of the VAE.

\subsection{BP based inference (VAE-BP-GMM)}
\label{sec:vae_bp_gmm}

In this second algorithm, we instead use BP as the iterative inference mechanism used to optimize $\hat{\bm{z}}$. BP minimizes the following loss function:

\begin{equation}
    \label{eq:bp_gmm}
    \mathcal{L}(\bm{x}, \hat{\bm{z}}) = \|\bm{f_\theta}(\hat{\bm{z}}) - \bm{x}\|^2 - \gamma \log p(\hat{\bm{z}}; \bm{M})
\end{equation}

where $\gamma$ is an hyperparameter weighting the influence of bottom-up and top-down mechanisms, and $\bm{f_\theta}$ denotes the decoder of the VAE. This gradient descent is parameterized by a learning rate $\lambda$. Once again, we can initialize the gradient descent using the estimate obtained with the encoder of the VAE. Derivations of this model are provided in appendix \ref{app:pc_model}, where the forward pass through the model is given in algorithm \ref{alg:bp_gmm}.

Investigating the relationship between PC and BP based inference, we have found that PC networks needed a larger number of iterations to convey information from the reconstruction error, and that this number of iterations was exponential with regard to the depth of the decoder. For this reason, experiments with the VAE-PC-GMM model were prohibitively slow, and we only conducted experiments with the remaining models.

\subsection{Restricting the iterative inference to the memory component (VAE-GMM)}
\label{sec:vae_gmm}

Backpropagating the reconstruction error might not be necessary when the estimate provided by the VAE encoder already conveys enough information from the observed input $\bm{x}$. As such, we also experiment with a simpler version of the previous model where iterative inference of $\hat{\bm{z}}$ only considers the top-down update rule coming from the memory. In this simpler version, the update rule for $\hat{\bm{z}}$ provided in algorithm \ref{alg:perceptual_inference} becomes (proof in appendix \ref{app:pc_model}):

\begin{equation}
\label{eq:vae_gmm}
    \hat{\bm{z}} \gets \text{softmax}\big(-\frac{\|\hat{\bm{z}} - \bm{M}\|_2^2}{2\sigma^2}\big) \cdot \bm{M}^\intercal
\end{equation}

This update mechanism is computationally lighter than the BP-based inference model, and has the advantage of being differentiable.

\subsection{Training the precision matrix (VAE-GMM*)}
\label{sec:vae_gmm_star}

For this last model, we suggest optimizing the precision coefficients $\bm{P} = \bm{\Sigma}^{-1}$ of the GMM. These coefficients condition the shape of the Gaussian mixtures. Some memory retrieval tasks might need to ignore partial information from the observed input $\bm{x}$ and as such could benefit from such an adaptation. In our experiments, we design such tasks on the CLEVR dataset, where the objective is to retrieve scenes using as input a shifted image of the same scene, or an image of the same scene where the colors of the objects have been modified. To properly accomplish these two tasks, the AM model needs to learn to give less importance to position information in the first case, and to color information in the second case.

The model starts from the estimate $\hat{\bm{z}}$ inferred by the VAE, and uses the following update rule:

\begin{equation}
    \hat{\bm{z}} \gets \text{softmax}\Big(-\frac{1}{2} (\hat{\bm{z}} - \bm{M})^\intercal \cdot \bm{P} \cdot (\hat{\bm{z}} - \bm{M})\Big) \cdot \bm{M}^\intercal
\end{equation}

During training, $\bm{P}$ is optimized using BP in order to reduce the mean squared error between the inferred representation $\hat{\bm{z}}$ and the correct memory pattern $\bm{M}_k^*$. During evaluation, the values of $\bm{P}$ are fixed.

\section{Experiments}
\label{sec:experiments}

In this section, we present the experiments performed to evaluate the proposed models. In all experiments, we measure performance using the percentage of properly retrieved patterns from associative memories containing $N=100$ patterns. We consider that a pattern is properly retrieved if the distance between the inferred representation $\hat{\bm{z}}$ and the correct memory pattern $\bm{M}_{k*}$ is lower than a threshold value chosen manually. All the implementation details (benchmark models, hyperparameters, training, dataset splits) are provided in appendix \ref{app:implementation_details}.

\subsection{Datasets}

We evaluate the proposed models on two image datasets: CIFAR10 and CLEVR. CIFAR10 \citep{Krizhevsky2009} (MIT License) consists of 32$\times$32 RGB images of 10 classes. We pretrain a VAE on the training set using a convolutional architecture, with a latent space of dimension $d=16$.

\begin{figure}[h]
    \centering
    \includegraphics[width=1.0\textwidth]{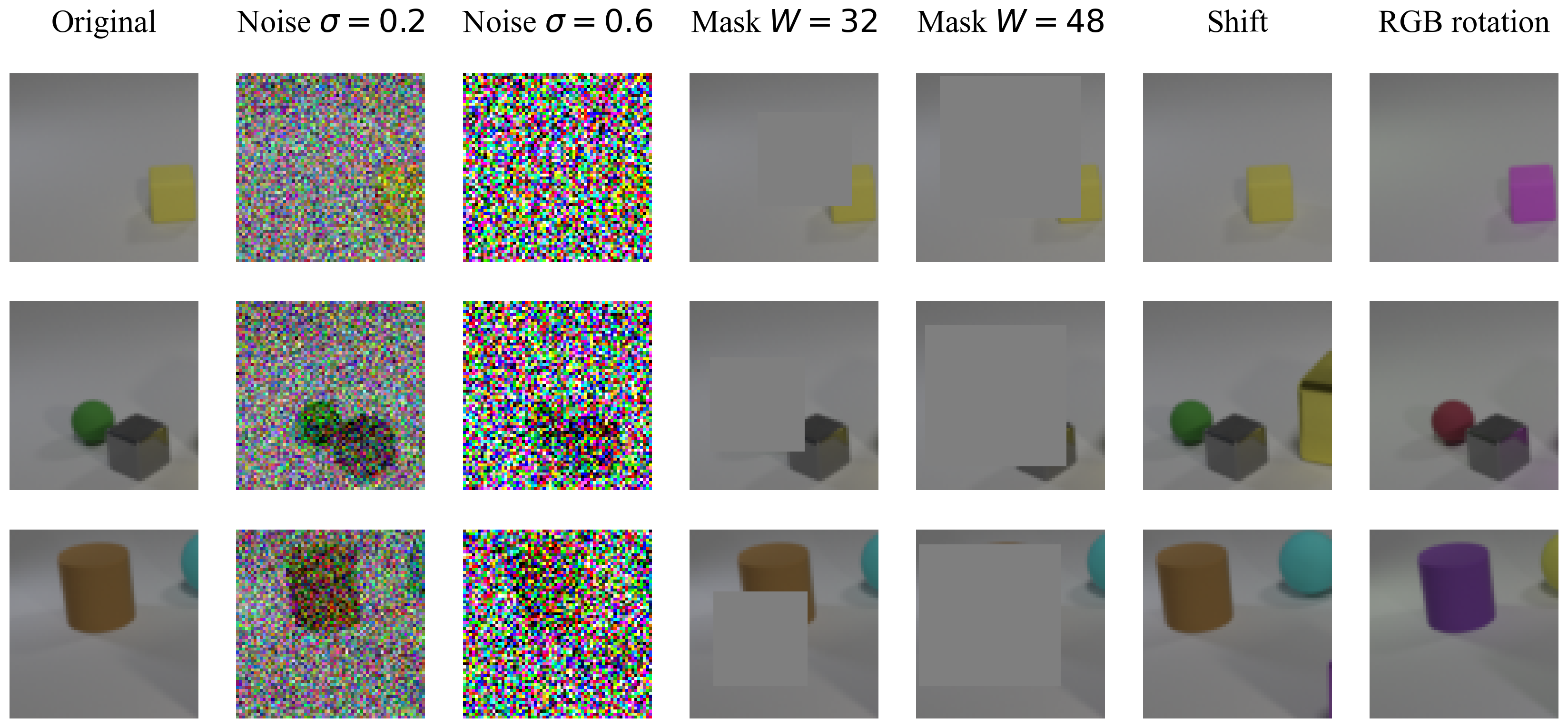}
    \caption{Examples of original, corrupted, and transformed images from the CLEVR dataset.}
    \label{fig:clevr_examples}
\end{figure}

The CLEVR dataset \cite{Johnson2017} (CC BY 4.0 License) consists of 64$\times$64 RGB images of 3D scenes composed of simple 3D objects. We use a pretrained MONet \citep{Burgess2019} model as the VAE, with a latent space of dimension $d=4\times16$. The MONet encoder infers a set representation given an input scene image. In our experiments, we flatten this set representation, which can lead to wrong measures of similarity between two representations. Indeed, a permuted set representation still encodes the same information, but once flattened the permuted and original representations might have a low similarity score. To limit this issue, we have capped the number of objects present in the scene images to 3, and capped the size of the set representation to 4. Examples of images from the CLEVR dataset along with possible transformations are presented in figure \ref{fig:clevr_examples}.

\subsection{Benchmark models}

We compare our AM models with Modern Continuous Hopfield Networks (MCHN), Neural Turing Machines (NTM) and End-to-End Memory Networks (MemN2N). For fair comparison, in all benchmark models, the memory is initialized with the stored patterns (we do not use the writing mechanisms of NTMs). We also experiment with these models in the representation space, where the memory is initialized with representations of the stored images, and the networks receive as input the representation of the corrupted image. We indicate this modification with the VAE prefix.

The NTM and MemN2N models are trained with BP using as loss function the mean squared error between the predicted output $\hat{\bm{x}}$ (respectively $\hat{\bm{z}}$ when using the VAE) and the correct memory pattern $\bm{M}_k^*$. For a fair comparison, this training is performed with clean inputs $\bm{x} = \bm{M}_k^*$, to avoid giving any prior information about the type of transformation the inputs might be corrupted with during evaluation. Finally, we also experiment with a version of the MCHN using a balanced GMM as starting prior distribution, that we denote GMM in our experiments. Equivalently, this is the version of our VAE-GMM algorithm without a representation component.

\subsection{Corrupted inputs}

First, we evaluate whether the associative memory models are able to properly retrieve stored patterns when given corrupted versions of these patterns. We apply different level of noise to the images ($\sigma \in \{0.2, 0.4, 0.6, \cdots, 2.0\}$), as well as different sizes of masks (16 and 24 pixels wide on the CIFAR10 dataset, and 32 and 48 pixels wide on the CLEVR dataset). The results of these experiments are reported in table \ref{tab:results_cifar10} and \ref{tab:results_clevr}. More figures and examples are provided in appendix \ref{app:additional_results}.

\begin{table}[h]
  \caption{Memory retrieval with corrupted inputs on the CIFAR10 dataset. Mean percentages of successful retrieval and standard deviations are reported.}
  \label{tab:results_cifar10}
  \centering
  \begin{tabular}{lllllllll}
    \toprule
    Input       & Clean input   & Noise $\sigma$=0.2 & Noise $\sigma$=0.6 & Mask W=16 & Mask W=24\\ 
    \midrule
    GMM         & \bm{$100\pm0.0$}   & \bm{$100\pm0.0$}   & $94.3\pm6.6$  & \bm{$99.7\pm1.3$}  & $48.7\pm12.7$ \\
    Hopfield    & $3.0\pm3.9$   & $3.9\pm4.0$   & $1.5\pm2.4$   & $2.3\pm3.6$   & $1.6\pm2.6$   \\
    NTM         & $99.3\pm2.4$  & $98.9\pm2.7$  & $91.6\pm15.0$ & $94.7\pm8.0$  & \bm{$62.7\pm20.9$}  \\
    MemN2N      & $81.1\pm11.3$   & $77.9\pm11.1$ & $65.8\pm13.4$ & $63.2\pm13.9$ & $32.3\pm11.5$   \\
    \midrule
    VAE-BP-GMM  & \bm{$100\pm0.0$}   & \bm{$100\pm0.0$}   & $81.1\pm8.9$  & $98.1\pm3.3$  & $39.5\pm10.7$  \\
    VAE-GMM     & \bm{$100\pm0.0$}   & \bm{$100\pm0.0$}   & $86.3\pm9.1$  & $98.3\pm2.8$  & $39.5\pm11.8$ \\
    VAE-Hopfield& $92.5\pm6.2$  & $91.7\pm6.0$  & $79.6\pm10.3$ & $82.5\pm8.1$  & $52.5\pm10.1$  \\
    VAE-NTM     & \bm{$100\pm0.0$}   & \bm{$100\pm0.0$}  &  \bm{$98.6\pm2.3$}  & $99.5\pm1.5$  & $58.7\pm10.2$ \\
    VAE-MemN2N  & $95.0\pm4.7$  & $92.6\pm6.1$  & $82.2\pm8.7$  & $82.2\pm8.0$  & $44.4\pm11.7$ \\
    \bottomrule
  \end{tabular}
\end{table}

\begin{table}[h]    
  \caption{Memory retrieval with corrupted inputs on the CLEVR dataset. Mean percentages of successful retrieval and standard deviations are reported.}
  \label{tab:results_clevr}
  \centering
  \begin{tabular}{lllllllll}
    \toprule
    Input       & Clean input   & Noise $\sigma$=0.2 & Noise $\sigma$=0.6 & Mask W=32 & Mask W=48\\ 
    \midrule
    GMM         & \bm{$100\pm0.0$}   & \bm{$100\pm0.0$}   & \bm{$73.0\pm10.1$} & $69.3\pm8.6$  & $30.3\pm10.9$ \\
    Hopfield    & $1.3\pm2.3$   & $1.3\pm2.8$   & $1.0\pm2.5$   & $0.8\pm1.8$   & $1.5\pm2.4$   \\
    NTM         & $82.1\pm13.1$ & $84.5\pm11.9$ & $47.8\pm16.8$  & $56.4\pm13.2$  & $17.6\pm9.7$  \\
    MemN2N      & $17\pm11.3$   & $16\pm12.3$   & $14.5\pm10.5$ & $9\pm11.2$    & $5.5\pm6.0$   \\
    \midrule
    VAE-BP-GMM  & \bm{$100\pm0.0$}   & $49.5\pm16.0$ & $2.0\pm2.5$   & $69.2\pm10.2$ & $36.5\pm5.8$  \\
    VAE-GMM     & \bm{$100\pm0.0$}   & $46.3\pm11.8$ & $4.4\pm4.4$   & $68.9\pm9.7$  & $31.5\pm11.6$ \\
    VAE-Hopfield& $94.2\pm4.9$  & $34.5\pm11.4$ & $2.7\pm4.0$   & $64.4\pm11.5$ & $32.7\pm9.9$  \\
    VAE-NTM     & $99.9\pm0.7$  & $40.8\pm10.0$ & $3.4\pm4.3$   & \bm{$75.0\pm10.2$} & \bm{$38.3\pm11.2$} \\
    VAE-MemN2N  & $97.9\pm3.1$  & $31.7\pm11.4$ & $2.4\pm3.1$   & $66.1\pm10.4$ & $34.2\pm10.1$ \\
    \bottomrule
  \end{tabular}
\end{table}

We can observe that the two models using dot-product attention, MCHN (Hopfield in the tables) and MemN2N, perform poorly compared to other methods, not even reaching 100\% of correctly retrieved patterns when presented with clean inputs. The NTM model, using cosine similarity based attention, works better and often outperforms the proposed models. Among the proposed models, we can see that the VAE-BP-GMM does not perform significantly better than the simpler VAE-GMM model.

An important observation from these results is that in most scenarios, the best performing methods do not use a representation component. In particular, the MONet model seems very sensitive to noise. However, our intuition is that memory retrieval on the representation level should be more powerful if the transformation applied on the observed inputs $\bm{x}$ has a limited effect on the representations.




\subsection{Scene transformations}

To verify this hypothesis, we perform two additional experiments on the CLEVR dataset. If we assume that the representations provided by MONet encode object positions, shapes and colors, then an AM model working on the representation level could perform well when we apply transformations on these specific object features. We propose two such scenarios. In the first scenario, we perform an RGB rotation on the images, which results in identical scenes with the exception of the object colors. In the second scenario, we take a shifted crop of the original CLEVR image, which can simulate a shifted point of view of the same scene. Examples images are displayed in figure \ref{fig:clevr_examples}.

We measure the success of memory retrieval with these transformed inputs using our models as well the NTM and MemN2N models. We also experiment with the VAE-GMM* model performing precision coefficient training. We expect the model to learn that some information conveyed by the encoder (for instance color in the first scenario) is irrelevant for memory retrieval. For fair comparison, we also experiment with the VAE-NTM and VAE-MemN2N models trained in the two scenarios, that we denote with an asterisk. The results are reported in table \ref{tab:results_vae}.

\begin{table}[h]
  \caption{Memory retrieval with transformed inputs on the CLEVR dataset. Mean percentages of successful retrieval and standard deviations are reported.}
  \label{tab:results_vae}
  \centering
  \begin{tabular}{lll}
    \toprule
    Input           & Color rotation        & Shift                 \\ 
    \midrule
    GMM             & $76\pm12.4$           & $28\pm8.5$            \\
    NTM             & $48.8\pm16.9$         & $0.7\pm1.9$           \\
    MemN2N          & $12.6\pm11.5$         & $6.4\pm8.1$           \\
    \midrule
    VAE-GMM         & $53.6\pm12.0$         & $63.1\pm10.2$         \\
    VAE-NTM         & $54.6\pm10.7$         & $61.5\pm11.6$         \\
    VAE-MemN2N      & $41.2\pm11.0$         & $56.2\pm12.1$         \\
    \midrule
    VAE-GMM*        & \bm{$98.9\pm2.5$}     & $93.7\pm5.3$          \\
    VAE-NTM*        & $98.2\pm3.2$          & \bm{$96.3\pm4.4$}     \\
    VAE-MemN2N*     & $91.6\pm6.9$          & $84.2\pm7.8$          \\
    \bottomrule
  \end{tabular}
\end{table}

Using a representation component improves the performance of the models in these scenarios. Additionally, training on the retrieval task where the transformations are applied provides almost perfect retrieval scores for the VAE-GMM and VAE-NTM models.

\section{Discussion}

We have proposed a formulation of auto-associative memories based on the FEP and the variational inference framework, where the stored patterns condition the prior probability on the representation $\bm{z}$. This framework has allowed us to draw a connection between MCHNs and the PC theory, as well as to design several AM models with high retrieval accuracy and robustness.

Combining representation and memory allows to retrieve patterns on the representation level, which decreases the size of the memory store, and performs retrieval based on a similarity measured in the latent space, where we can expect more meaningful features to appear. For instance, we have shown that this allows to recognize visual scenes from a shifted point of view, while pixel-level AM models failed most of the time. Our results also seem to demonstrate that distance based attention as used in our models (see equation \ref{eq:vae_gmm}) outperforms dot-product based attention as used in MCHNs, although this performance gap could be mitigated with normalization techniques, as hinted by the results obtained with NTMs, that use cosine similarity instead of dot-product.

The proposed PC and BP based models take into account information from the query $\bm{x}$ at each inference iteration, by directly minimizing an energy function that depends on $\bm{x}$ (eq. \ref{eq:vfe_pc_expression} and eq. \ref{eq:bp_gmm}). In contrast, the Kanerva Machine \citep{wu2018kanerva, wu2018learning} or the VAE-GMM and VAE-GMM* models we proposed only use this information to output an initial estimate $\hat{\bm{z}}$ that is later optimized in order to minimize an energy function only depending on the memory. According to our experiments, this feature of the PC and BP based models does not seem useful for memory retrieval. However, in other applicative settings, these models could benefit from their ability to weight query pattern information and stored patterns information, performing a form of memory-aided iterative perceptual inference.

The proposed formulation is limited to content-based reading mechanisms. In contrast, models such as NTMs offer a larger variety of addressing schemes as well as writing mechanisms that allow to only store relevant information in the memory. Another limitation of this work is that we did not investigate the memory capacity and convergence properties of the proposed models. Future work should focus on these analyses, and possible improvements of the PC-based inference, for instance using precision weighting, or the "learning to optimize" approach of \citep{marino2018iterative}.

\section{Acknowledgements}

This work was funded by the CY Cergy-Paris University Foundation (Facebook grant) and partially by Labex MME-DII, France (ANR11-LBX-0023-01).

\bibliography{references}


\newpage

\section*{Checklist}

\begin{enumerate}

\item For all authors...
\begin{enumerate}
  \item Do the main claims made in the abstract and introduction accurately reflect the paper's contributions and scope?
    \answerYes{We believe that the claimed contributions in Introduction are properly reflected in sections \ref{sec:theory} and \ref{sec:methods}.}
  \item Did you describe the limitations of your work?
    \answerYes{See the last paragraph of the Discussion.}
  \item Did you discuss any potential negative societal impacts of your work?
    \answerNo{We did not identify any direct potential negative societal impact. If any indirect potential impact is brought to our attention, we would be happy to discuss it.}
  \item Have you read the ethics review guidelines and ensured that your paper conforms to them?
    \answerYes{}
\end{enumerate}

\item If you are including theoretical results...
\begin{enumerate}
  \item Did you state the full set of assumptions of all theoretical results?
    \answerYes{The assumptions and definitions used in our derivations are stated (see sections \ref{sec:pc_mchn} and \ref{sec:methods}, and appendices \ref{app:fe_derivations},\ref{app:mchn_derivations} and \ref{app:pc_model}).}
        \item Did you include complete proofs of all theoretical results?
    \answerYes{We provide detailed derivations of our models and theoretical results in appendices \ref{app:fe_derivations}, \ref{app:mchn_derivations} and \ref{app:pc_model}.}
\end{enumerate}

\item If you ran experiments...
\begin{enumerate}
  \item Did you include the code, data, and instructions needed to reproduce the main experimental results (either in the supplemental material or as a URL)?
    \answerYes{See appendix \ref{app:implementation_details}.}
  \item Did you specify all the training details (e.g., data splits, hyperparameters, how they were chosen)?
    \answerYes{See appendix \ref{app:implementation_details}.}
    \item Did you report error bars (e.g., with respect to the random seed after running experiments multiple times)?
    \answerYes{See tables \ref{tab:results_cifar10}, \ref{tab:results_clevr} and \ref{tab:results_vae}, as well as figures in appendix \ref{app:additional_results}.}
    \item Did you include the total amount of compute and the type of resources used (e.g., type of GPUs, internal cluster, or cloud provider)?
    \answerYes{See appendix \ref{app:implementation_details}.}
\end{enumerate}

\item If you are using existing assets (e.g., code, data, models) or curating/releasing new assets...
\begin{enumerate}
  \item If your work uses existing assets, did you cite the creators?
    \answerYes{Datasets were credited in section \ref{sec:experiments}, code assets are credited in appendix \ref{app:implementation_details}.}
  \item Did you mention the license of the assets?
    \answerYes{Licences for the datasets were mentionned in section \ref{sec:experiments}, the license for used code assets is mentionned in \ref{app:implementation_details}.}
  \item Did you include any new assets either in the supplemental material or as a URL?
    \answerYes{Implementations of our models are linked in appendix \ref{app:implementation_details}.}
  \item Did you discuss whether and how consent was obtained from people whose data you're using/curating?
    \answerNA{}
  \item Did you discuss whether the data you are using/curating contains personally identifiable information or offensive content?
    \answerNo{The data we used are commonly used ML datasets. CLEVR is an automatically generated dataset and to our knowledge, the CIFAR10 dataset does not contain any personally identifiable or offensive content.}
\end{enumerate}

\item If you used crowdsourcing or conducted research with human subjects...
\begin{enumerate}
  \item Did you include the full text of instructions given to participants and screenshots, if applicable?
    \answerNA{}
  \item Did you describe any potential participant risks, with links to Institutional Review Board (IRB) approvals, if applicable?
    \answerNA{}
  \item Did you include the estimated hourly wage paid to participants and the total amount spent on participant compensation?
    \answerNA{}
\end{enumerate}

\end{enumerate}


\newpage

\appendix

\section{General VFE derivation}
\label{app:fe_derivations}

In this section, we provide the derivations for the VFE expression given in equation \ref{eq:vfe_pc_expression}.

The FEP formulation of PC is based on several assumptions that allow to derive a simpler expression of the VFE. For simplicity, we provide a derivation for the case where $L=1$, meaning that there is only one layer in the generative model of $\bm{x}$ based on $\bm{z}$. At the end of the derivation, we provide the expression in the general case. We start from a different equivalent formulation of the VFE:

\begin{equation}
    F(\bm{x}) = \int_{\bm{z}} E(\bm{x}, \bm{z}) q(\bm{z}) \bm{dz} + \int_{\bm{z}} \log\big(q(\bm{z})\big) q(\bm{z}) \bm{dz}
\end{equation}

where $E(\bm{x}, \bm{z}) = -\log p(\bm{x}, \bm{z})$ is called the energy. We assume that the distribution $q(\bm{z})$ takes a Gaussian form $q(\bm{z}) = \mathcal{N}(\bm{z}; \hat{\bm{z}}, \zeta\mathbb{I})$. The mean of this approximate posterior, $\hat{\bm{z}}$, corresponds to the inferred representation being optimized by the PC networks. Integrating this definition into the VFE expression, we obtain:

\begin{equation}
    F(\bm{x})= - \frac{d}{2} \log(2 \pi \zeta) - \frac{d}{2} + \int_{\bm{z}}E(\bm{x}, \bm{z})q(\bm{z}) \bm{dz}
\end{equation}

We assume that the approximate posterior is tightly shape around its mean $\hat{\bm{z}}$, allowing us to use the Taylor expansion of $E(\bm{x}, \bm{z})$ around this value:

\begin{equation}
    E(\bm{x}, \bm{z}) \approx E(\bm{x}, \hat{\bm{z}}) +  \big( \nabla_{\hat{\bm{z}}} E(\bm{x}, \hat{\bm{z}}) \big) \cdot (\bm{z} - \hat{\bm{z}})
\end{equation}

We can now derive an expression of the VFE that depends on $\hat{\bm{z}}$ and not longer involves integrals:

\begin{align}
    F(\bm{x}, \hat{\bm{z}}) &\approx E(\bm{x}, \hat{\bm{z}}) + \big( \nabla_{\hat{\bm{z}}} E(\bm{x}, \hat{\bm{z}}) \big) \cdot \int_{\bm{z}} (\bm{z} - \hat{\bm{z}}) q(\bm{z}) \bm{dz}+ C \\
    &\approx  E(\bm{x}, \hat{\bm{z}}) + C \\
    &\approx -\log p(\bm{x}|\hat{\bm{z}}) - \log p(\hat{\bm{z}}; \bm{M}) + C
\end{align}

where C is a quantity that does not depend on $\bm{x}$ and $\hat{\bm{z}}$. Finally, generalizing this expression to $L$ layers, and assuming that each layer in the generative model takes the form of a Gaussian distribution with mean $\bm{f_\theta}^l(\hat{\bm{h}_{l+1}})$ and variance $\mathbb{I}$, we obtain:

\begin{equation}
    \begin{aligned}
        F(\bm{x}, \hat{\bm{h}_1}, \dots, \hat{\bm{h}_{L-1}}, \hat{\bm{z}}) = &\sum_{l=0}^{L-1} \frac{1}{2} \| \hat{\bm{h}_l} - \bm{f_\theta}^{l}(\hat{\bm{h}_{l+1}}) \|^2 \\
        &- \log p(\hat{\bm{z}};\bm{M}) \\
        &+ C'
    \end{aligned}
\end{equation}

where $C'$ includes other terms independent from $\{\bm{x}, \hat{\bm{h}_1}, \cdots, \hat{\bm{z}}\}$ coming from the derivation of the logarithms of the multivariate Gaussians. More detailed derivations can be found in \citep{Buckley2017}, without the memory dependency, but this has virtually no impact on the derivations.

\newpage
\section{MCHN derivations}
\label{app:mchn_derivations}

In this section, we provide the derivations for the expression of the VFE and the update rule for MCHNs (equations \ref{eq:vfe_hopfield} and \ref{eq:vfe_hopfield_update} in the main text). Using the assumptions listed in section \ref{sec:pc_mchn}, we can derive an expression of the VFE that closely resembles the energy function proposed in the MCHN paper \cite{Ramsauer2020}:

\begin{align}
    F(\hat{\bm{z}}) &= - \log p (\hat{\bm{z}}; \bm{M}) + C \\
    &= - \log \sum_{k=1}^N \frac{\exp \{\frac{\beta}{2}\bm{M_k}^\intercal \cdot \bm{M_k}\}}{\sum_{k'=1}^N \exp \{\frac{\beta}{2}\bm{M_{k'}}^\intercal \cdot \bm{M_{k'}}\}} \frac{1}{\sqrt{2 \pi \beta^{-d}}} \exp \{ -\frac{\beta}{2}(\hat{\bm{z}} - \bm{M_k})^\intercal \cdot (\hat{\bm{z}} - \bm{M_k})\} + C \\
    &= - \log \sum_{k=1}^{N} \exp \{\frac{\beta}{2}\bm{M_k}^\intercal \cdot \bm{M_k}\} \exp \{ -\frac{\beta}{2}(\hat{\bm{z}} - \bm{M_k})^\intercal \cdot (\hat{\bm{z}} - \bm{M_k})\} + C' \\
    &= - \log \sum_{k=1}^N \exp\{ \beta \hat{\bm{z}}^\intercal \cdot \bm{M_k} \} \exp \{-\frac{\beta}{2} \hat{\bm{z}}^\intercal \cdot \hat{\bm{z}}\} + C' \\
    &= \frac{\beta}{2} \hat{\bm{z}}^\intercal \cdot \hat{\bm{z}} - \log \sum_{k=1}^N \exp \{  \beta \hat{\bm{z}}^\intercal \cdot \bm{M_k} \} + C'
\end{align}

Up to an additive constant and a factor $\beta$, this expression is equivalent to the energy function proposed in \citep{Ramsauer2020}. According to the FEP formulation of PC, the neural dynamics performing iterative optimization of $\hat{\bm{z}}$ can be derived from the gradient descent update with regard to the VFE. We start by deriving this gradient:

\begin{align}
    \nabla_{\hat{\bm{z}}}F(\hat{\bm{z}}) &= \beta \hat{\bm{z}} - \frac{\sum_{k=1}^N \exp \{\beta \hat{\bm{z}}^\intercal \cdot \bm{M_k}\} \cdot (\beta \bm{M_k})}{\sum_{k'=1}^N \exp \{\beta \hat{\bm{z}}^\intercal \cdot \bm{M_{k'}}\}} \\
    &= \beta \{ \hat{\bm{z}} - \text{softmax}(\beta \hat{\bm{z}}^\intercal \cdot \bm{M}) \bm{M}^\intercal \}
\end{align}

Which yields the following update rule for $\hat{\bm{z}}$:

\begin{equation}
    \hat{\bm{z}} \gets \hat{\bm{z}} + \alpha \beta \big\{ \text{softmax}(\beta \hat{\bm{z}}^\intercal \cdot \bm{M}) \bm{M}^\intercal - \hat{\bm{z}} \big\}
\end{equation}

where $\alpha$ is the rate of the gradient descent. In particular, when $\alpha = \frac{1}{\beta}$ we obtain the update rule of the MCHN:

\begin{equation}
    \hat{\bm{z}} \gets \text{softmax}(\beta \hat{\bm{z}}^\intercal \cdot \bm{M}) \bm{M}^\intercal
\end{equation}

\begin{figure}[ht]
    \begin{subfigure}{0.49\textwidth}
        \centering
        \includegraphics[width=\textwidth]{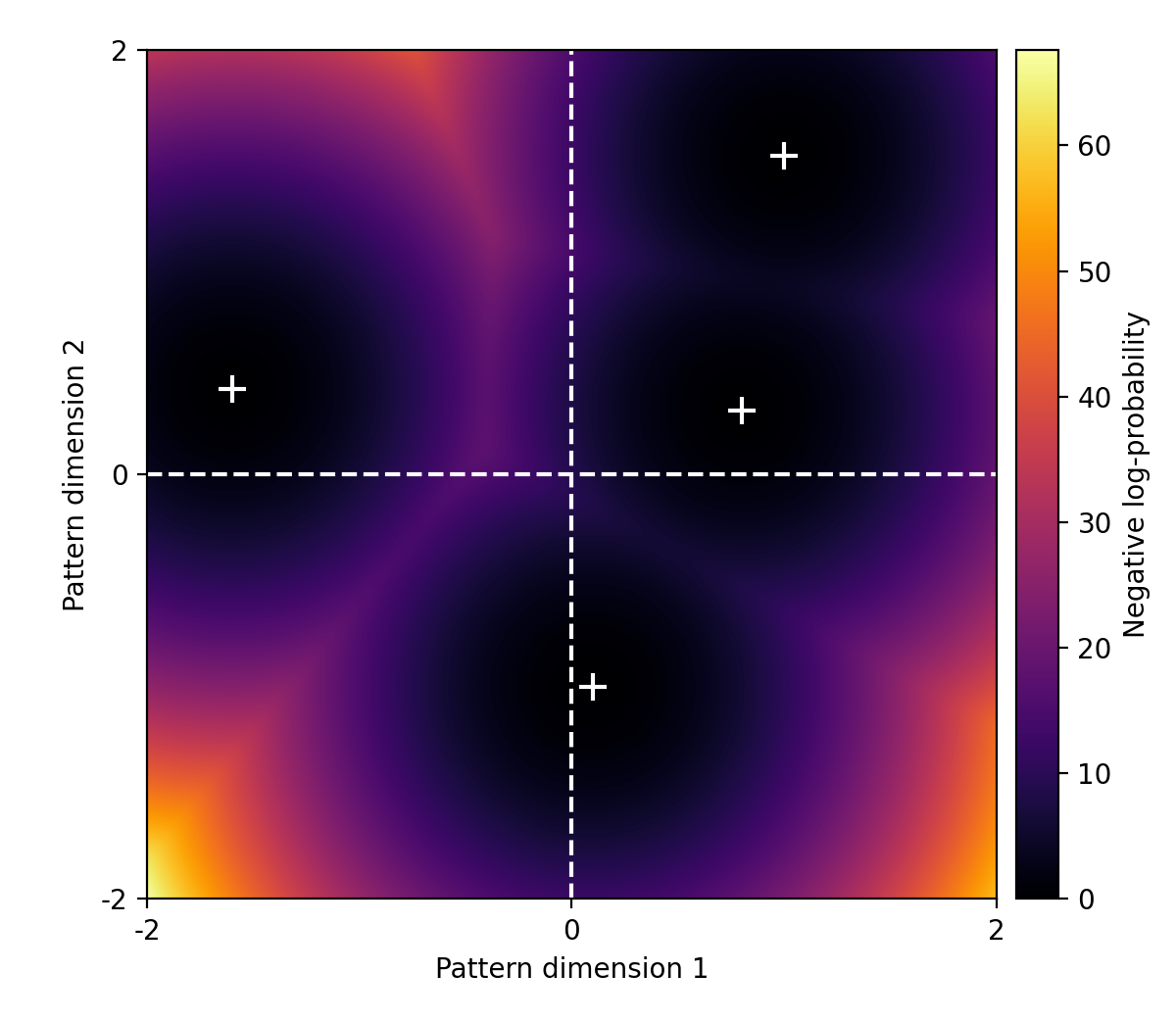}
    \end{subfigure}
    ~
    \begin{subfigure}{0.49\textwidth}
        \centering
        \includegraphics[width=\columnwidth]{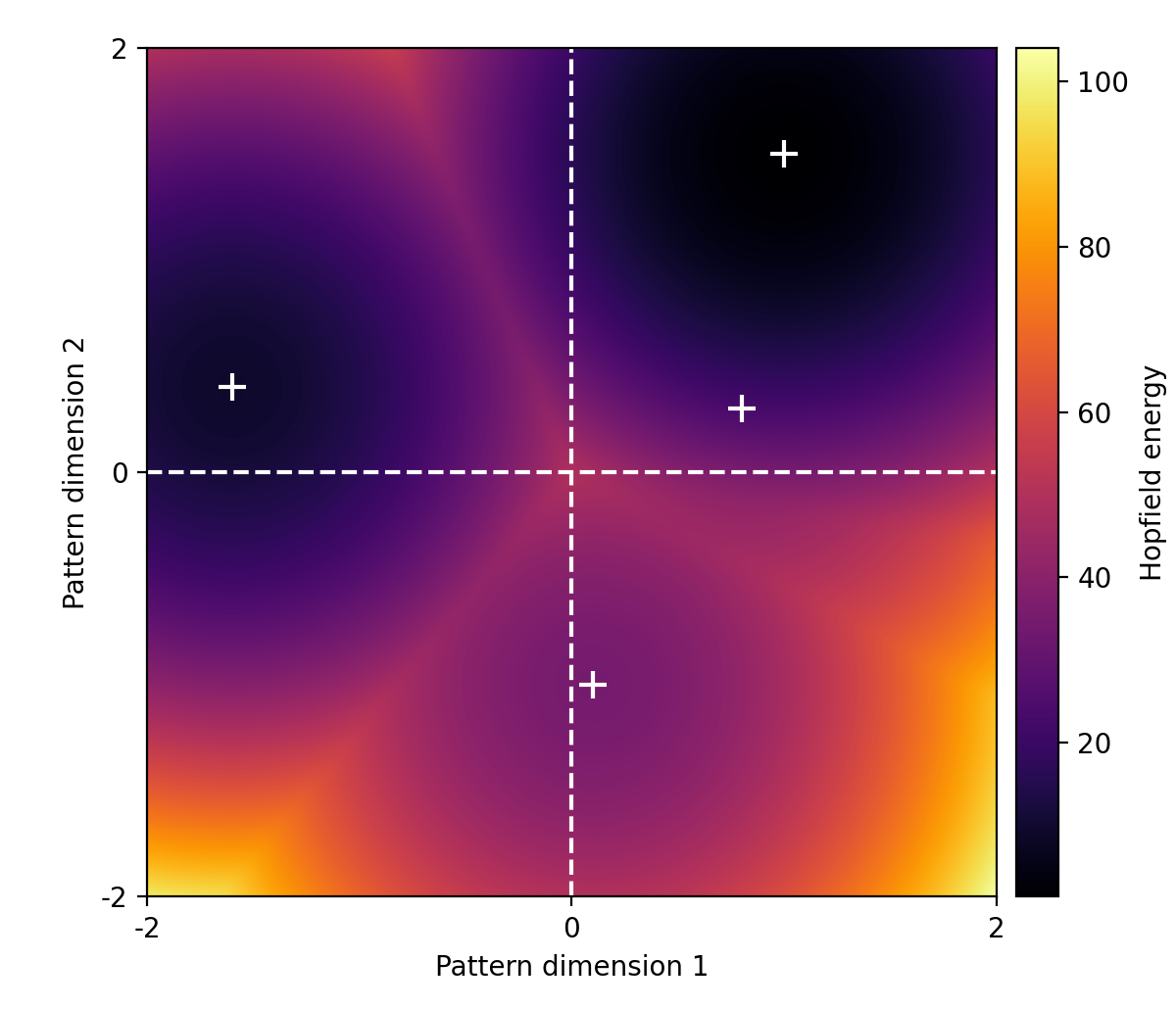}
    \end{subfigure}
    \caption{Illustration of the VFE landscape induced by the balanced (ours, left) and biased (MCHN, right) GMM prior distribution, for an AM of four 2D patterns represented in white. These figures were obtained with $\beta=100$ and $\sigma=0.2$.}
    \label{fig:energy_functions}
\end{figure}

Looking back at the prior distribution $p(\bm{z};\bm{M})$, the GMM is biased towards representations of larger Euclidean norm. This means that stored patterns $\bm{M}_k$ aligned with other patterns of larger norms cannot attract the dynamics of the MCHN. This is represented in figure \ref{fig:energy_functions} where we have displayed two energy landscapes for an AM containing four 2D patterns. The VFE computed with a balanced GMM model (left) comprises a local minimum for each pattern, which is not the case for the VFE computed with the MCHN prior distribution.

\newpage
\section{Derivations of the proposed models}
\label{app:pc_model}

\subsection{(VAE-)PC-GMM derivations}

In this section, we provide derivations for the PC-GMM (and VAE-PC-GMM when $\hat{\bm{z}}$ is initialized with the VAE encoder). We start by expressing the VFE using the GMM prior distribution:

\begin{equation}
    \begin{aligned}
        F(\bm{x}, \hat{\bm{h}_1}, \dots, \hat{\bm{h}_{L-1}}, \hat{\bm{z}}) = &\sum_{l=0}^{L-1} \frac{1}{2} \| \hat{\bm{h}_l} - \bm{f_\theta}^{l}(\hat{\bm{h}_{l+1}}) \|^2 \\
        &- \log \sum_{k=1}^N \exp \{ -\frac{1}{2\sigma^2}(\hat{\bm{z}} - \bm{M_k})^\intercal \cdot (\hat{\bm{z}} - \bm{M_k})\} \\
        &+ C'
    \end{aligned}
\end{equation}

where the constant C' contains other terms coming from the GMM expression that do not depend on $\{\bm{x}, \hat{\bm{h}_1}, \dots, \hat{\bm{h}_{L-1}}, \hat{\bm{z}}\}$. According to the FEP formulation of PC, the neural dynamics simulate a gradient descent on this energy function. We can thus derive the update rules for the approximate posterior means $\{\hat{\bm{h}_1}, \dots, \hat{\bm{h}_{L-1}}, \hat{\bm{z}}\}$. For all $1\leq l \leq L$:

\begin{align}
    \hat{\bm{h}_l} \gets \hat{\bm{h}_l} - \alpha \nabla_{\hat{h}_l} F(\bm{x}, \hat{\bm{h}_1}, \dots, \hat{\bm{h}_{L-1}}, \hat{\bm{z}})
\end{align}

where $\alpha$ is the rate of the gradient descent. For the intermediate layers$1\leq l < L$, we obtain the following update rule:

\begin{equation}
    \hat{\bm{h}_l} \gets \hat{\bm{h}_l} - \underbrace{\alpha \big(\hat{\bm{h}_l} - \bm{f_\theta}^l(\hat{\bm{h}_{l+1}})\big)}_{\text{Top-down}} + \underbrace{\alpha \bm{f_\theta}^{l-1'}(\hat{\bm{h}_l}) \cdot \big(\hat{\bm{h}_{l-1}} - \bm{f_\theta}^{l-1}(\hat{\bm{h}_l})\big)}_{\text{Bottom-up}}
\end{equation}

This update rule combines top-down information pulling $\hat{\bm{h}_l}$ towards its prediction coming from the upper layer, and bottom-up information pulling it towards a value that reduces the prediction error on the lower layer. It is useful to introduce the notation $\bm{\epsilon}_l = \hat{\bm{h}_l} - \bm{f_\theta}^l(\hat{\bm{h}_{l+1}})$ called the prediction error on layer $l$. In the PC theory, at each layer a population of neurons encodes this quantity, while another encodes the current estimate $\hat{\bm{h}}_l$. For the last layer, the bottom-up signal is identical, but the top-down signal pulls $\hat{\bm{z}}$ towards values that maximize the prior $p(\bm{z})$:

\begin{equation}
    \hat{\bm{z}} \gets \hat{\bm{z}} + \underbrace{\frac{\alpha}{\sigma^2} \Big(
        \text{softmax}\big(-\frac{\|\hat{\bm{z}} - \bm{M}\|_2^2}{2\sigma^2}\big) \cdot \bm{M}^\intercal - \hat{\bm{z}} \Big)}_{\text{Top-down}} + \underbrace{\alpha \big(\bm{f_\theta}^{L-1'}(\hat{\bm{z}}) \cdot \bm{\epsilon}_{L-1} \big)}_{\text{Bottom-up}}
\end{equation}

\begin{figure*}[ht]
    \centering
    \includegraphics[width=\textwidth]{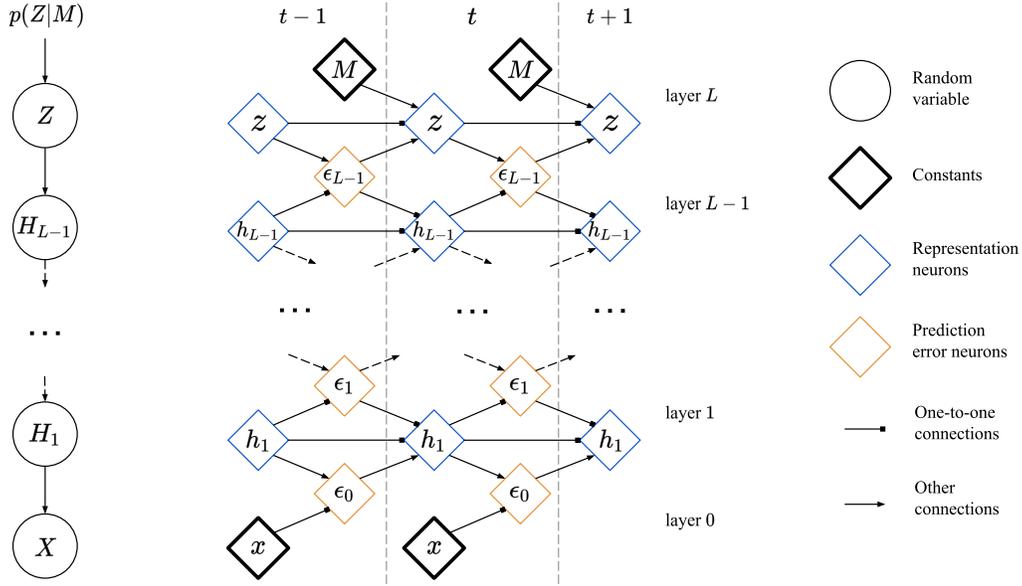}
    \caption{Left: probabilistic graphical model of our system. Right: corresponding PC network.}
    \label{fig:pc_model}
\end{figure*}

These update rules can be applied iteratively, which results in a dynamical system viewed in the PC theory as an RNN. Figure \ref{fig:pc_model} represents this RNN unfolded in time (right) along with the assumed hierarchical probabilistic model (left). Algorithm \ref{alg:perceptual_inference} describes the forward pass through this PC network.

\begin{algorithm*}
    \DontPrintSemicolon
    \textbf{Parameters:} $\bm{\theta}, \alpha, \sigma, \bm{M}$ \\
    \textbf{Input:} $\bm{x}$ \\
    
    \SetAlgoLined
    Initialize $\big\{\hat{\bm{z}}, \hat{\bm{h}_{L-1}}, \dots,  \hat{\bm{h}_1} \big\}$ \;
    \For{$0 \leq t < T$}{
        \tcc{Compute prediction errors}
        \For{$0 \leq l < L$}{
            $\bm{\epsilon}_l \gets \hat{\bm{h}_l} - \bm{f_\theta}^l(\hat{\bm{h}_{l+1}}) $\;
        }
        \tcc{Update hidden representations}
        \For{$1 \leq l < L$}{
            $\hat{\bm{h}_l} \gets \hat{\bm{h}_l} + \alpha \big(\bm{f_\theta}^{l-1'}(\hat{\bm{h}_l}) \cdot \bm{\epsilon}_{l-1}- \bm{\epsilon}_l \big)$ \;
        }
        \tcc{Update z}
        $\hat{\bm{z}} \gets \hat{\bm{z}} + \alpha \big(\bm{f_\theta}^{L-1'}(\hat{\bm{z}}) \cdot \bm{\epsilon}_{L-1} \big) + \frac{\alpha}{\sigma^2} \Big(
        \text{softmax}\big(-\frac{\|\hat{\bm{z}} - \bm{M}\|_2^2}{2\sigma^2}\big) \cdot \bm{M}^\intercal - \hat{\bm{z}} \Big)$ \;
    }
    \caption{PC-GMM Memory retrieval}
    \label{alg:perceptual_inference}
\end{algorithm*}

\subsection{(VAE-)BP-GMM derivations}

Here we show that the PC-GMM dynamics approximate the gradient descent updates resulting from the application of BP on the loss function of the BP-GMM model. To obtain this result, we must assume that the predictions $\bm{f_\theta}^l(\hat{\bm{h}_{l+1}})$ remain constant during this iterative inference process. This hypothesis, often called "fixed prediction assumption", is required to prove this results. Note that in our PC-GMM algorithm, we have updated the predictions at each iteration, so the result proven here do not apply in our case. Still this result is interesting as it highlights the relationship between PC-based inference and BP-based inference. Therefore, we assume here that the predictions $\bm{f_\theta}^{l-1}(\hat{\bm{h}_l})$ and the derivatives $\bm{f_\theta}^{l-1'}(\hat{\bm{h}_l})$ are fixed during the inference process. Only the prediction errors $\bm{\epsilon}_l$ and the approximate posterior means $\hat{\bm{h}_l}$ are updated. We recall that the loss function used in the BP-GMM model is defined as:

\begin{equation}
    \mathcal{L}(\bm{x}, \hat{\bm{z}}) = \|\bm{f_\theta}(\hat{\bm{z}}) - \bm{x}\|^2 - \gamma \log p(\hat{\bm{z}}; \bm{M})
\end{equation}

Given an input $\bm{x}$ and a memory matrix $\bm{M}$, the dynamics of the PC network will reach equilibrium when for all $l$, $\nabla_{\hat{\bm{h}_l}} F = 0$. For the intermediate layers, this is verified when:

\begin{equation}
\label{eq:recurrence_epsilon}
    \bm{\epsilon}_l = \bm{f_\theta}^{l-1'}(\hat{\bm{h}_l}) \cdot \bm{\epsilon}_{l-1}
\end{equation}

Equivalently, we can derive the expression of the gradients provided by BP on the intermediate quantities $\hat{\bm{h}_l}$. On the bottom layer, we have:

\begin{align}
    \nabla_{\hat{\bm{h}_1}} \| \bm{f_\theta}(\hat{\bm{z}}) - \bm{x} \|_2^2 &= \nabla_{\hat{\bm{h}_1}} \| \bm{f_\theta}^0(\hat{\bm{h}_1}) - \bm{x} \|_2^2 \\
    &= 2\bm{f_\theta}^{0'}(\hat{\bm{h}_1}) \cdot (\bm{f_\theta}^0(\hat{\bm{h}_1}) - \bm{x}) \\
    &= -2 \bm{f_\theta}^{0'}(\hat{\bm{h}_1}) \cdot \bm{\epsilon_0} \\
    &= -2 \bm{\epsilon}_1
\end{align}

Using the chain rule, we can derive the gradient with regard to $\hat{\bm{h}_l}$ based on the gradient with regard to $\hat{\bm{h}_{l-1}}$. We observe that we obtain the same recurrence relation between gradients $\nabla_{\hat{\bm{h}_l}}\| \bm{f_\theta}(\hat{\bm{z}}) - \bm{x} \|_2^2$ than the one we obtained with prediction errors $\bm{\epsilon}_l$ at equilibrium (equation \ref{eq:recurrence_epsilon}):

\begin{equation}
    \nabla_{\hat{\bm{h}_l}}\| \bm{f_\theta}(\hat{\bm{z}}) - \bm{x} \|_2^2 = \bm{f_\theta}^{l-1'}(\hat{\bm{h}_l}) \cdot \nabla_{\hat{\bm{h}_{l-1}}}\| \bm{f_\theta}(\hat{\bm{z}}) - \bm{x} \|_2^2
\end{equation}

Therefore, according to the induction principle, we can conclude that for all layers $1\leq l<L$:

\begin{equation}
    \bm{\epsilon}_l = -2 \nabla_{\hat{\bm{h}_l}} \|\bm{f_\theta}(\hat{\bm{z}}) - \bm{x} \|_2^2
\end{equation}

Now, looking at the topmost layer, we can compare the update rule for $\hat{\bm{z}}$ prescribed by BP and PC. For PC, we have seen that the update rule is:

\begin{equation}
    \hat{\bm{z}} \gets \hat{\bm{z}} + \underbrace{\frac{\alpha}{\sigma^2} \Big(
        \text{softmax}\big(-\frac{\|\hat{\bm{z}} - \bm{M}\|_2^2}{2\sigma^2}\big) \cdot \bm{M}^\intercal - \hat{\bm{z}} \Big)}_{\text{Top-down}} + \underbrace{\alpha \big(\bm{f_\theta}^{L-1'}(\hat{\bm{z}}) \cdot \bm{\epsilon}_{L-1} \big)}_{\text{Bottom-up}}
\end{equation}

For BP, we once again use the chain rule:

\begin{align}
    \nabla_{\hat{\bm{z}}} \mathcal{L} &= \nabla_{\hat{\bm{z}}} \big( -\gamma \log p(\hat{\bm{z}}; \bm{M})\big)  + \bm{f_\theta}^{L-1'}(\hat{\bm{z}}) \cdot \nabla_{\hat{\bm{h}_{L-1}}} \|\bm{f_\theta}(\hat{\bm{z}}) - \bm{x} \|_2^2 \\
    &= \underbrace{-\frac{\gamma}{\sigma^2} \Big( \text{softmax}\big(-\frac{\|\hat{\bm{z}} - \bm{M}\|_2^2}{2\sigma^2}\big) \cdot \bm{M}^\intercal - \hat{\bm{z}} \Big)}_{\text{Top-down}} - \underbrace{2\bm{f_\theta}^{L-1'}(\hat{\bm{z}}) \bm{\epsilon_{L-1}}}_{\text{Bottom-up}}
\end{align}

Taking $\gamma=\alpha=2$ yields the exact same iterative inference update rule for both approaches. If we remove the "fixed prediction assumption" this equivalence no longer stands. However, this proves that the two approaches are closely related. In practice, we found that the two models performed similarly but that the PC-GMM approach was prohibitively slow to propagate information for very deep generative models.

The iterative algorithm is described in algorithm \ref{alg:bp_gmm}.

\begin{algorithm*}
    \DontPrintSemicolon
    \textbf{Parameters:} $\bm{\theta}, \alpha, \sigma, \gamma, \bm{M}$ \\
    \textbf{Input:} $\bm{x}$ \\
    
    \SetAlgoLined
    Initialize $\hat{\bm{z}}$ \;
    \For{$0 \leq t < T$}{
        \tcc{Compute the prediction}
        $\hat{\bm{x}} \gets \bm{f_\theta}(\hat{\bm{z}})$ \;
        
        \tcc{Compute the energy function}
        $\mathcal{L} \gets \|\hat{\bm{x}}  - \bm{x}\|^2 - \gamma \log p(\hat{\bm{z}}; \bm{M})$ \;

        \tcc{Update z using BP}
        $\hat{\bm{z}} \gets \hat{\bm{z}} - \alpha \nabla_{\hat{\bm{z}}} \mathcal{L}$ \;
    }
    \caption{BP-GMM Memory retrieval}
    \label{alg:bp_gmm}
\end{algorithm*}

\subsection{VAE-GMM derivations}

The derivation of the VAE-GMM model is straightforward. We simply remove the bottom-up information pathway of the VAE-PC-GMM model and instead consider that the amortized inference performed by the encoder already conveys the necessary information from $\bm{x}$. The update rule for $\hat{\bm{z}}$ becomes:

\begin{equation}
\label{eq:vae_gmm_smooth}
    \hat{\bm{z}} \gets \hat{\bm{z}} + \frac{\alpha}{\sigma^2} \Big(\text{softmax}\big(-\frac{\|\hat{\bm{z}} - \bm{M}\|_2^2}{2\sigma^2}\big) \cdot \bm{M}^\intercal - \hat{\bm{z}} \Big)
\end{equation}

If we choose $\alpha = \sigma^2$, we obtain the update rule of the VAE-GMM model:

\begin{equation}
    \hat{\bm{z}} \gets\text{softmax}\big(-\frac{\|\hat{\bm{z}} - \bm{M}\|_2^2}{2\sigma^2}\big) \cdot \bm{M}^\intercal
\end{equation}

\section{Implementation details}
\label{app:implementation_details}

All the presented experiments were performed on a single NVIDIA GeForce GTX 1060 GPU. 

Training was performed on the training sets of the two datasets, and the results reported in this article were obtained on the testing sets. 

The training hyperparameters (learning rate, number of steps for the MemN2N model) were optimized in order to achieve the lowest prediction error on the training set. The memory retrieval hyperparameters ($\sigma$ for GMM models, $\beta$ for MCHN models, $\gamma$ for the BP-GMM model) were optimized in order to achieve the highest successful retrieval percentage on the training set.

The reported results were obtained using one seed for the VAE, 5 seeds for the AM models that require training (MemN2N, NTM, VAE-GMM*), and 10 seeds for the memory retrieval scenarios that include randomness (noise and mask).

We provide the code including the implementation of the proposed models, our implementation of the benchmark models, the different memory retrieval scenarios and the hyperparameter values we have experimented with: \url{https://github.com/sino7/predictive_coding_associative_memories}.

Our implementation of the MONet model was adapted from the implementation provided in the github repository \url{https://github.com/baudm/MONet-pytorch}, and we used the provided pretrained weights on the CLEVR dataset.

\newpage
\section{Additional results}
\label{app:additional_results}

\subsection{Ablation study}

In this section, we investigate the impact of two features of our model: the initialization of $\hat{\bm{z}}$ using the encoder, and the use of a balanced GMM instead of the biased GMM of the MCHN model (see appendix \ref{app:mchn_derivations}).

We have reproduced the memory retrieval experiment with noisy inputs on two new model variations: BP-GMM (without VAE initialization) and VAE-BP-Hopfield. The VAE-BP-Hopfield is the biased GMM version of the VAE-BP-GMM model, where the BP is used to perform iterative inference in order to minimize the loss function:

\begin{equation}
    \mathcal{L} = \|\bm{f_\theta}(\hat{\bm{z}}) - \bm{x}\|^2 + \gamma \big(\frac{1}{2} \hat{\bm{z}}^\intercal \cdot \hat{\bm{z}} -\frac{1}{\beta}\log \sum_{k=1}^N \exp\{\beta \hat{\bm{z}}^\intercal\cdot \bm{M}_k\} \big)
\end{equation}

\begin{figure}[ht]
    \begin{subfigure}{0.49\textwidth}
        \centering
        \includegraphics[width=\textwidth]{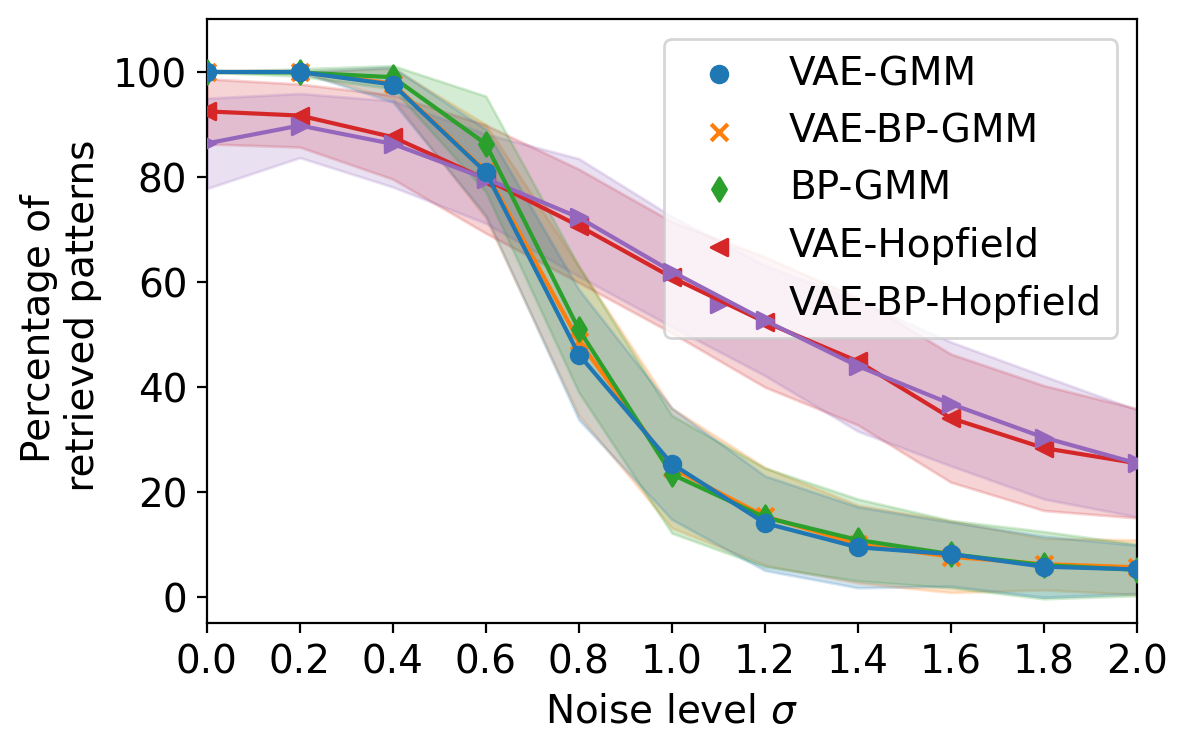}
        \caption{CIFAR10}
    \end{subfigure}
    ~
    \begin{subfigure}{0.49\textwidth}
        \centering
        \includegraphics[width=\columnwidth]{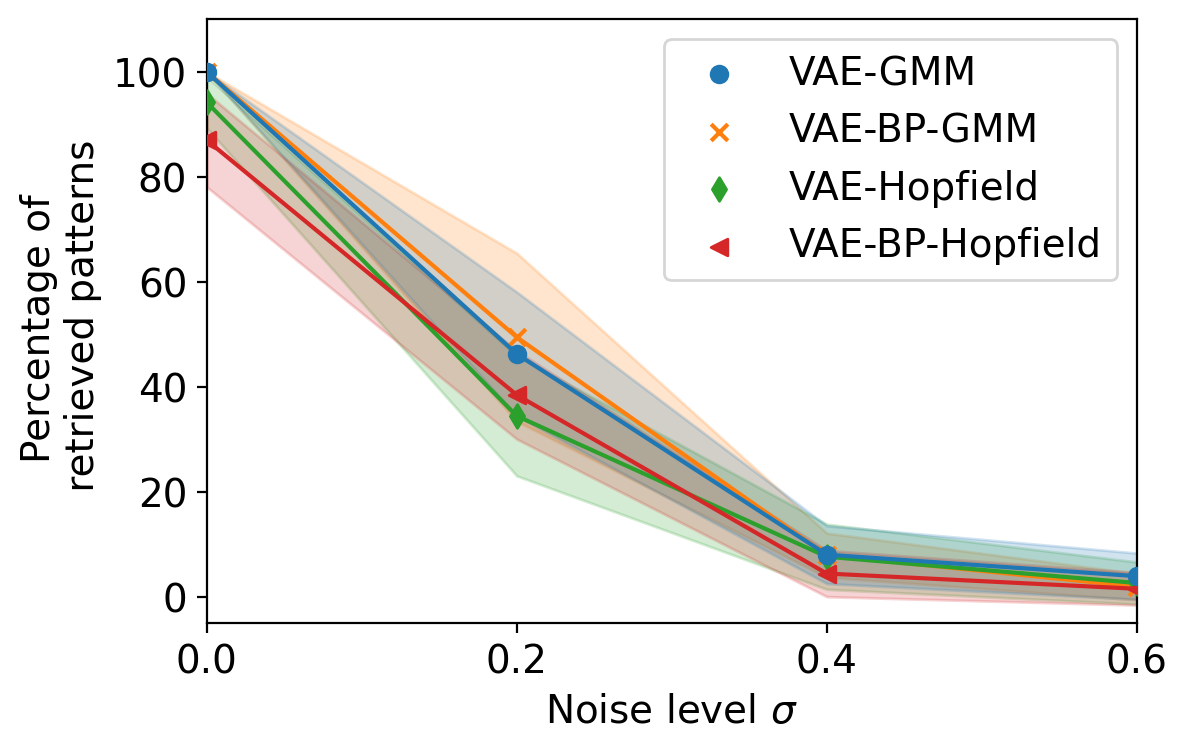}
        \caption{CLEVR}
    \end{subfigure}
    \caption{Percentage of successful memory retrieval using inputs corrupted with a noise of varying standard deviation $\sigma$. Intervals indicate standard deviation.}
    \label{fig:ablation_noise}
\end{figure}

We report in figure \ref{fig:ablation_noise} the percentage of successful memory retrieval using these different models. On the CLEVR dataset, the BP-GMM (without VAE initialization) always failed to retrieve the correct memory pattern. On the CIFAR10 dataset, it performed exactly the same as the VAE-GMM and VAE-BP-GMM models. This argues in favor of the simpler VAE-GMM model, that seems to convey information from the input $\bm{x}$ properly enough. This is also observed by comparing the results using the VAE-BP-Hopfield model and the simpler VAE-Hopfield model.

We can note that on the CIFAR10 dataset, Hopfield based retrieval is more robust to very high levels of noise. We believe that in this case, the bias towards patterns of high L2 norm might partially counter the indirect effect of the noise onto $\hat{\bm{z}}$. On the CLEVR dataset this synergy is not observed and all models rapidly fail to retrieve patterns in memory.

On the other hand, we can see that Hopfield-based models never reach perfect retrieval even when the presented inputs are clean, which argues in favor of the balanced GMM alternative.

\subsection{Analysis of the trained VAE-GMM* precision coefficients}

In this section, we investigate the effect of precision coefficient learning in the VAE-GMM* model. After training in two different scenarios: RGB rotation and shift, we compare the learned precision coefficients. For a more straightforward analysis, we have restricted the precision matrix to be diagonal. This way we can directly identify dimensions of the representation $\bm{z}$ that are deemed more or less relevant for memory retrieval in both scenarios. 

\begin{figure}[ht]
    \centering
    \includegraphics[width=\textwidth]{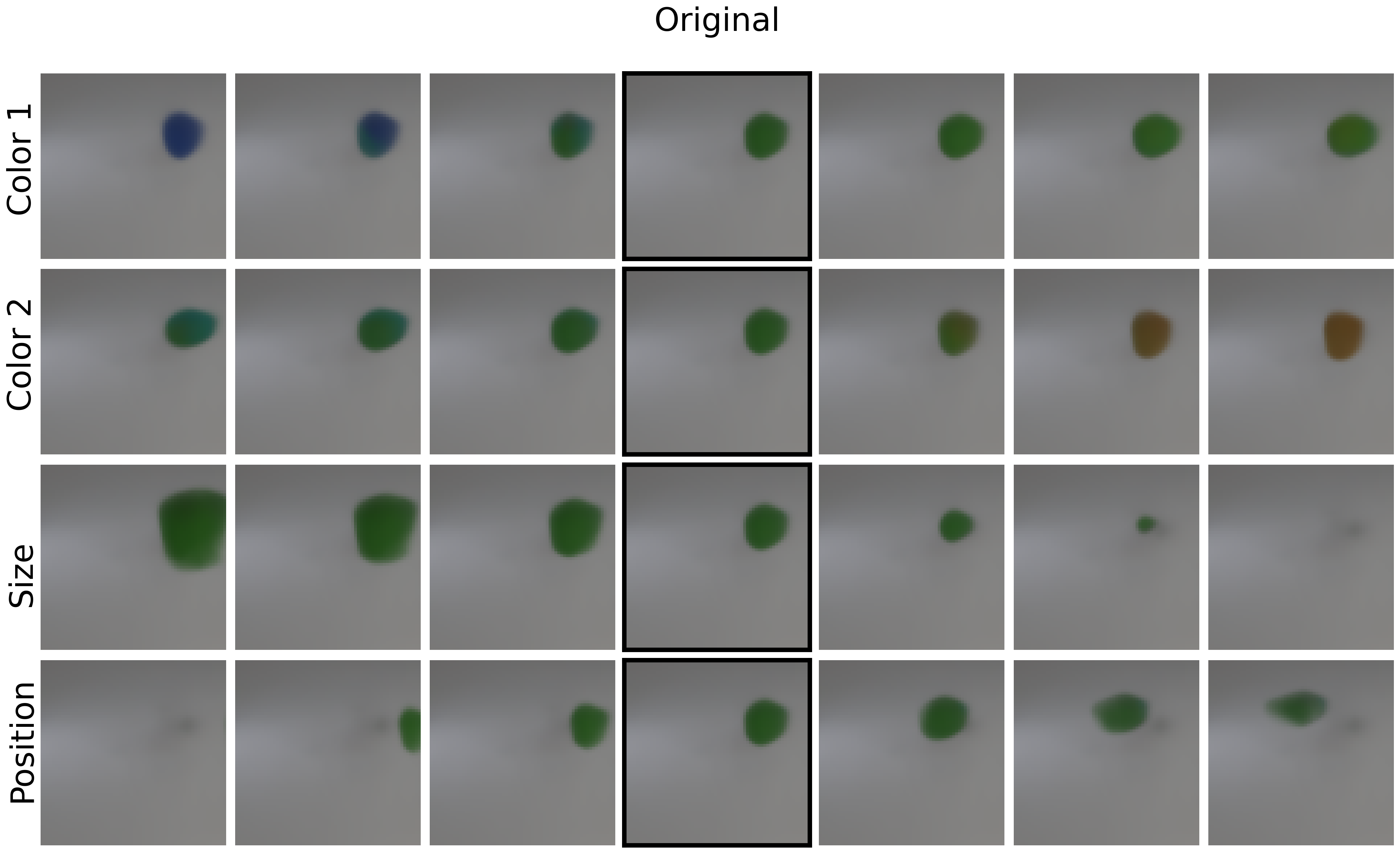}
    \caption{Varying the representation along some dimensions. The dimensions for the first two rows seem to encode object color. The dimension for row 3 seems to encode object size. The dimension for row 4 seems to encode object horizontal position.}
    \label{fig:clevr_dimensions}
\end{figure}

The CLEVR representation is structured into four object representations (one for the background and the three others for possible objects in the scene). We have identified four dimensions of the object representation where the precision coefficients in both scenarios presented the highest disagreement. We have then sampled an image from the CLEVR dataset and made variations along these dimensions to observe their effect on the decoded images. As shown in figure \ref{fig:clevr_dimensions}, these four dimensions can be interpreted as encoding color, size and position.

\begin{figure}[ht]
    \centering
    \includegraphics[width=\textwidth]{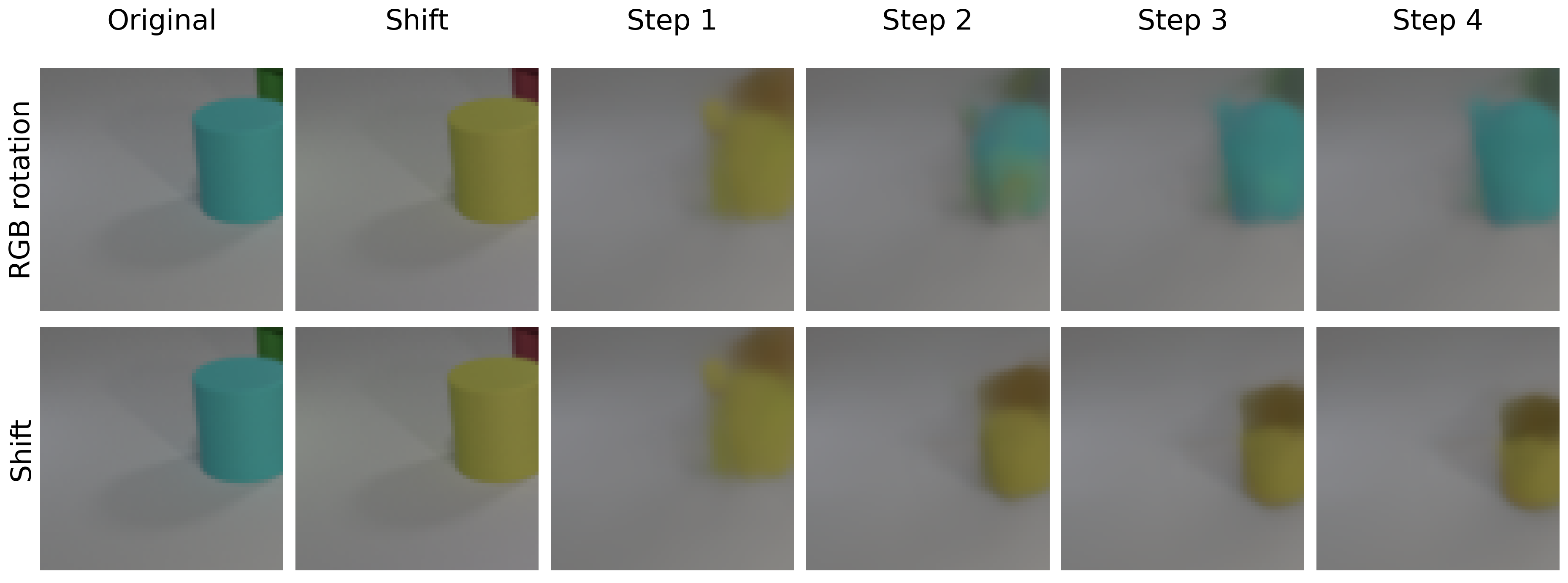}
    \caption{Images decoded from the representations $\hat{\bm{z}}$ at different steps of the inference process, using the VAE-GMM* models trained with the RGB rotation (first row) and with shifted images (second row). For this experiment, we have use the VAE-GMM model with a lower update coefficient $\alpha$ (see equation \ref{eq:vae_gmm_smooth}) to observe a smooth convergence.}
    \label{fig:clevr_comparison_color}
\end{figure}

For the VAE-GMM* model trained with the RGB rotation, the precision coefficients corresponding to the color dimension were lower, meaning that the retrieval mechanism gave less importance to these features when comparing the inferred representation $\hat{\bm{z}}$ with the memory patterns $\bm{M}_k$. Conversely, for the VAE-GMM* model trained with shifted images, the precision coefficients corresponding to the position and size (to a lower a extent) were lower. Consequently, the two trained models (with the same memory content) can react differently to the same input pattern, as shown in figure \ref{fig:clevr_comparison_color}.

We can observe that for an input image obtained by applying an RGB rotation on one of the stored patterns, the model trained on the correct task properly retrieves the pattern, while the model trained with shifted images instead converges to a stored pattern corresponding to a similar object in a different position. These results confirm our intuition that adaptation of the precision coefficients can help the proposed VAE-GMM model to give more or less importance to certain representation features for memory retrieval.

\subsection{One-shot generation}

\begin{figure}[ht]
    \centering
    \includegraphics[width=\textwidth]{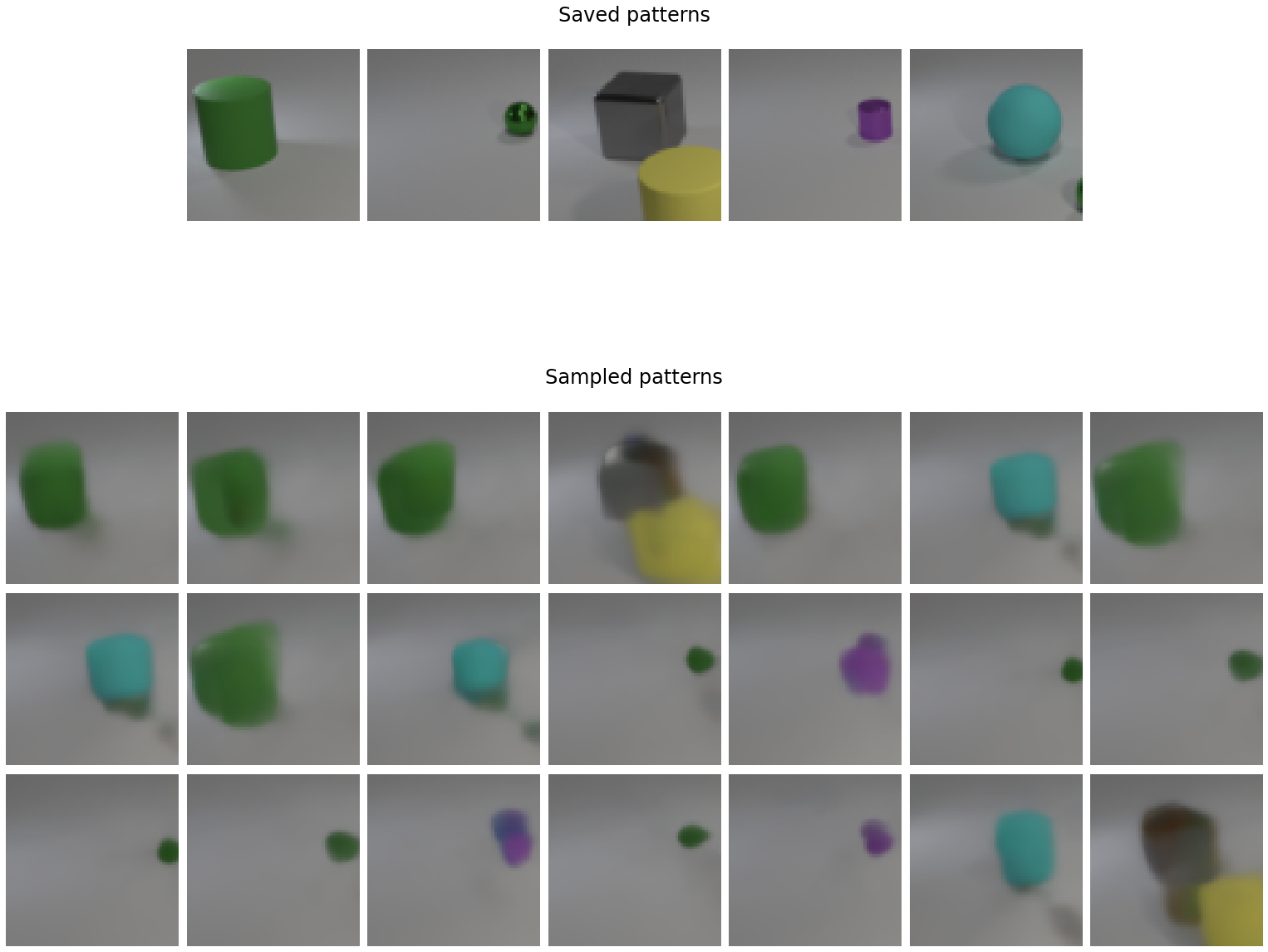}
    \caption{Examples of generated images.}
    \label{fig:one_shot_generation}
\end{figure}

In this section, we display examples of images sampled with our memory-dependent generative model on the CLEVR dataset. The first row of figure \ref{fig:one_shot_generation} contains the input patterns written in the memory. The writing operation simply consists in encoding the images and building the memory matrix with the obtained column vectors. In the bottom of this figure are images sampled from the memory-dependent generative model. We can observe that the sampled images contain similar objects in the same positions, with slight variations of shape, size, position or color.

\subsection{Capacity}

We have tried measuring the retrieval success rate with a varying number of patterns in the memory. The models based on a balanced GMM can achieve 100\% of successful retrievals when no noise is applied, for a better comparison, we thus experiment with input patterns corrupted with a noise of standard deviation $\sigma=0.6$. We compare the performance of the GMM models with or without the representation component (respectively VAE-GMM and GMM) as well as the performance with the MCHN variants (VAE-Hopfield). Since the MCHN applied on the raw pixel level scores very low in our initial experiments (see table \ref{tab:results_cifar10} and table \ref{tab:results_clevr}), we only experiment here with the variant working on the representation level. This experiment is conducted on the CIFAR10 dataset, with memory stores of size varying from $N=5$ to $N=10000$. 

\begin{figure}[ht]
    \centering
    \includegraphics[width=0.5\textwidth]{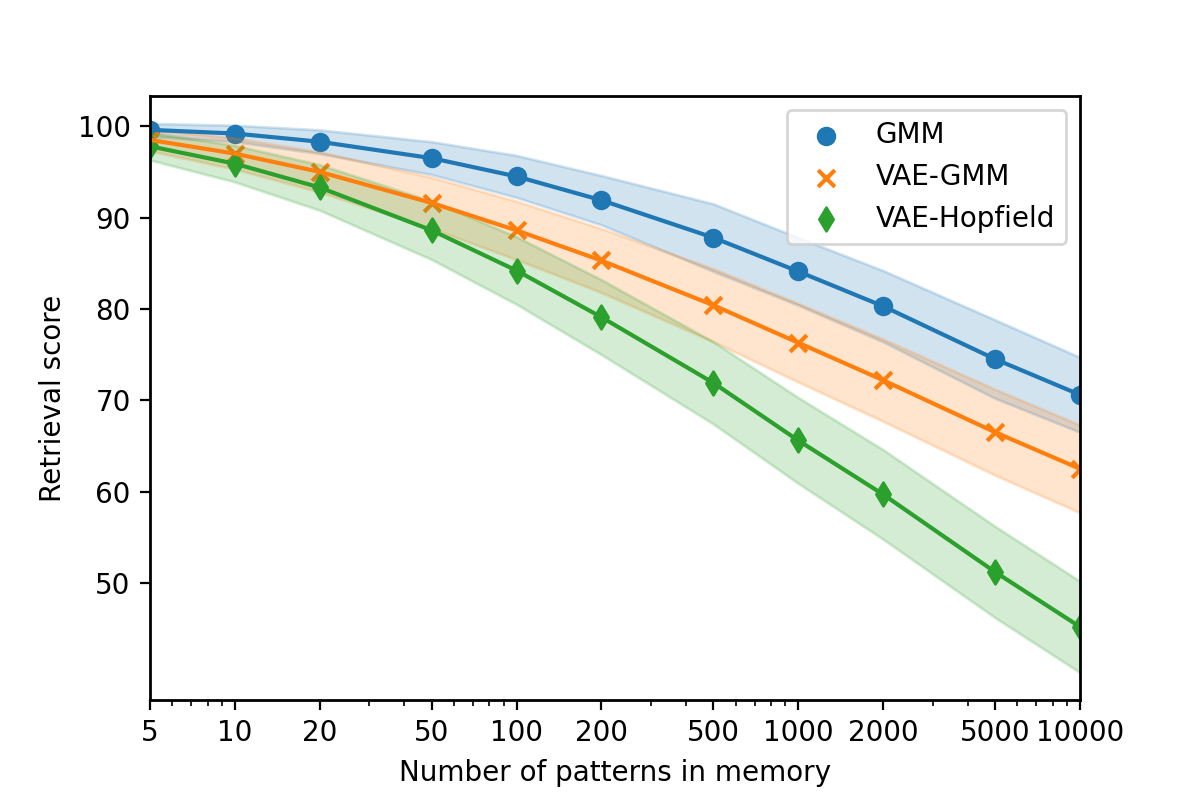}
    \caption{Evolution of the retrieval rate according to the number of memory patterns.}
    \label{fig:capacity_xp}
\end{figure}

The results are displayed in figure \ref{fig:capacity_xp}. We can observe that the performance of the VAE-Hopfield model drops faster than the performance of the models based on the balanced GMM implementation we proposed. Another result is that the use of a representation component does not seem to improve the capacity of the model.

\subsection{Additional figures}

\begin{figure}[ht]
    \centering
    \includegraphics[width=0.8\textwidth]{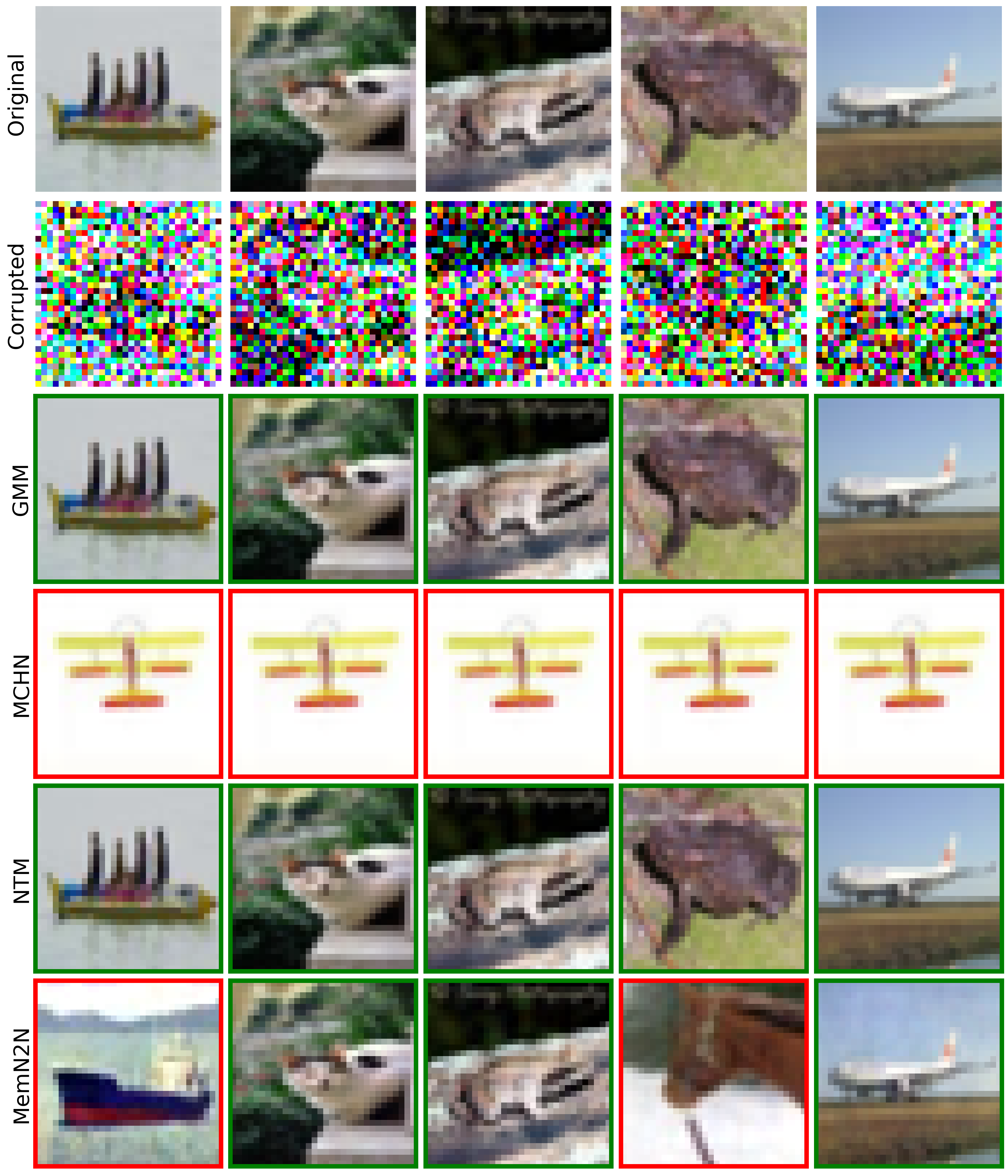}
    \caption{Retrieved images using inputs corrupted with a noise of standard deviation $\sigma=0.6$.}
    \label{fig:cifar10_noise_comparison_raw}
\end{figure}

\begin{figure}[ht]
    \centering
    \includegraphics[width=0.9\textwidth]{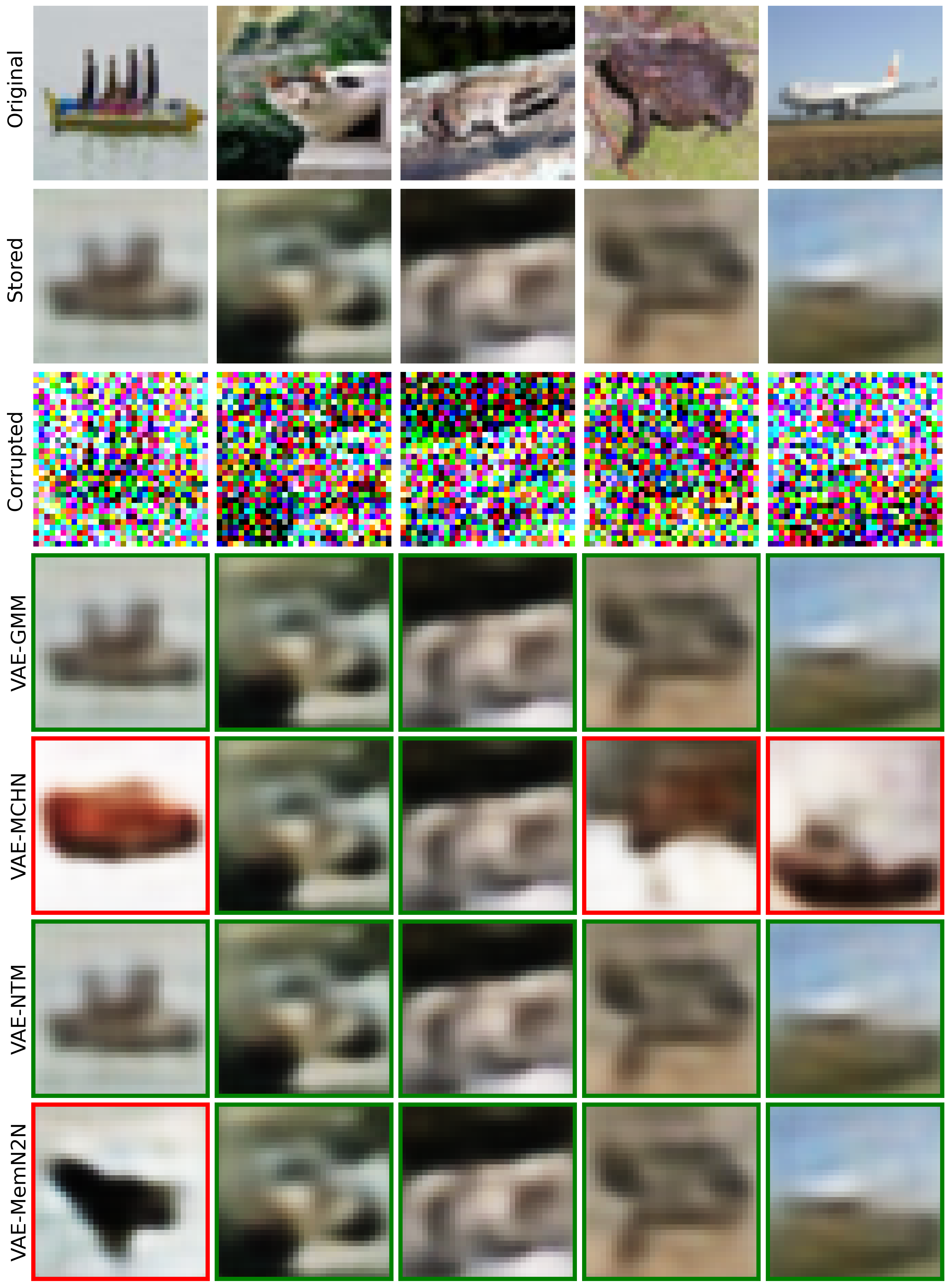}
    \caption{Examples of retrieved images with different models, using inputs corrupted with a noise of standard deviation $\sigma=0.6$.}
    \label{fig:cifar10_noise_comparison_vae}
\end{figure}

\begin{figure}[ht]
    \centering
    \includegraphics[width=0.9\textwidth]{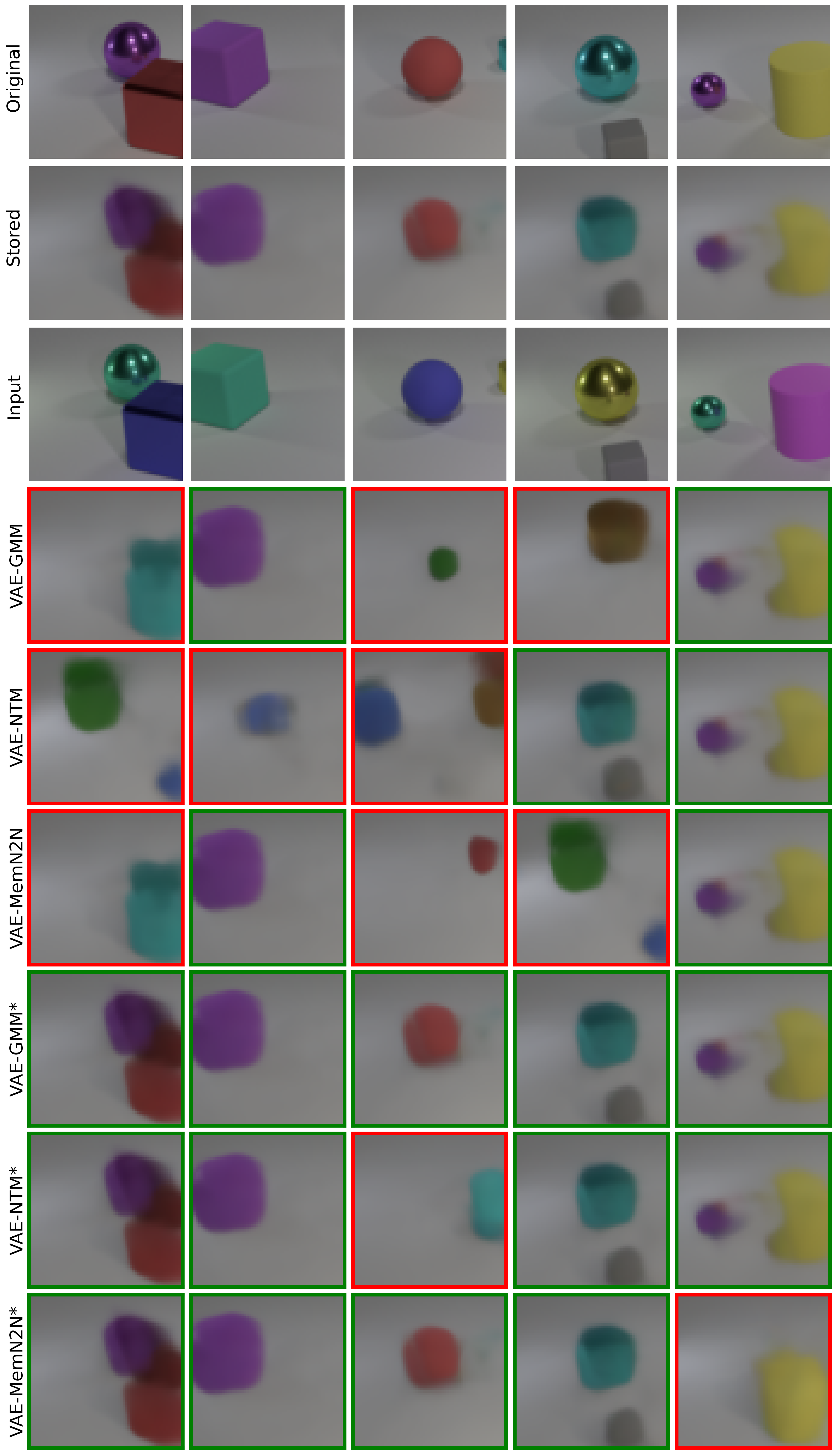}
    \caption{Retrieved images with different models in the RGB rotation scenario.}
    \label{fig:clevr_color_comparison}
\end{figure}

In this section, we provide examples of input and retrieved images. On the CIFAR10 dataset, we provide examples using AM models working on the pixel level in figure \ref{fig:cifar10_noise_comparison_raw} and AM models working on the representation level in figure \ref{fig:cifar10_noise_comparison_vae}. On the CLEVR dataset, we provide examples using models working on the representation level, with and without dedicated training in the "RGB rotation" scenario, in figure \ref{fig:clevr_color_comparison}.

\end{document}